\documentclass{article} 
\usepackage{iclr2025_conference,times}
\iclrfinalcopy

\usepackage{amsthm}
\usepackage{natbib}
\usepackage{pgf}
\usepackage{svg}
\usepackage{wrapfig}
\usepackage{tabularray}
\usepackage{physics}
\usepackage{amsmath}
\usepackage{xspace}

\newtheorem{lemma}{Lemma}
\newtheorem{theorem}{Theorem}
\newtheorem{proposition}{Proposition}
\newtheorem{corollary}{Corollary}

\newcommand{\mathset}[1]{\left\{ #1 \right\}}

\newcommand{\quantile}[2]{\sQ\left(#1;#2\right)}
\newcommand{\invquantile}[2]{\sQ^{-1}\left(#1;#2\right)}

\newcommand{\parbold}[1]{\textbf{#1.}\xspace}
\usepackage{booktabs} 

\usepackage{amsmath,amsfonts,bm}

\def\eqref#1{equation~\ref{#1}}

\def\1{\bm{1}}

\def\vh{{\bm{h}}}

\def\vs{{\bm{s}}}
\def\vt{{\bm{t}}}
\def\vu{{\bm{u}}}
\def\vv{{\bm{v}}}

\def\vx{{\bm{x}}}

\def\vz{{\bm{z}}}
\def\vepsilon{\bm{\epsilon}}

\def\mI{{\bm{I}}}

\DeclareMathAlphabet{\mathsfit}{\encodingdefault}{\sfdefault}{m}{sl}
\SetMathAlphabet{\mathsfit}{bold}{\encodingdefault}{\sfdefault}{bx}{n}

\def\gB{{\mathcal{B}}}
\def\gC{{\mathcal{C}}}
\def\gD{{\mathcal{D}}}

\def\gG{{\mathcal{G}}}
\def\gH{{\mathcal{H}}}

\def\gN{{\mathcal{N}}}

\def\gR{{\mathcal{R}}}

\def\gU{{\mathcal{U}}}

\def\gX{{\mathcal{X}}}
\def\gY{{\mathcal{Y}}}

\def\sI{{\mathbb{I}}}

\def\sQ{{\mathbb{Q}}}
\def\sR{{\mathbb{R}}}

\newcommand{\E}{\mathbb{E}}

\newcommand{\pertx}{\tilde{\vx}}

\newcommand{\subref}[2]{\autoref{#1}-#2}

\newcommand{\method}{BinCP\xspace}
\newcommand{\testf}{\mathrm{accept}}

\usepackage[ruled,vlined]{algorithm2e}
\usepackage{booktabs}
\usepackage{multirow}

\usepackage{hyperref}
\usepackage{url}

\usepackage[inline]{enumitem}

\title{Robust Conformal Prediction with a Single Binary Certificate}

\author{Soroush H. Zargarbashi \\
CISPA Helmholtz Center for Information Security\\
\texttt{zargarbashi@cs.uni-koeln.de} \\
\And
Aleksandar Bojchevski \\
University of Cologne \\
\texttt{bojchevski@cs.uni-koeln.de} \\
}

\newcommand{\dcal}{\gD_\mathrm{cal}}
\newcommand{\xtest}{\vx_{n+1}}
\newcommand{\xtestpert}{\tilde\vx_{n+1}}
\newcommand{\lowerval}{\mathrm{c}^\downarrow}
\newcommand{\upperval}{\mathrm{c}^\uparrow}

\begin{document}

\maketitle

\begin{abstract}
Conformal prediction (CP) converts any model's output to prediction sets with a guarantee to cover the true label with (adjustable) high probability. Robust CP extends this guarantee to worst-case (adversarial) inputs. Existing baselines achieve robustness by bounding randomly smoothed conformity scores. In practice, they need expensive Monte-Carlo (MC) sampling (e.g. $\sim10^4$ samples per point) to maintain an acceptable set size. We propose a robust conformal prediction that produces smaller sets even with significantly lower MC samples (e.g. 150 for CIFAR10).
Our approach binarizes samples with 
an adjustable (or automatically adjusted) threshold
selected to preserve the coverage guarantee. Remarkably, we prove that robustness can be achieved by computing \emph{only one} binary certificate, unlike previous methods that certify each calibration (or test) point. Thus, our method is faster and returns smaller robust sets. We also eliminate a previous limitation that requires a bounded score function.
\end{abstract}

\section{Introduction}
\label{sec:intro}

Despite their extensive applications, modern neural networks lack reliability as their output probability estimates are uncalibrated \citep{Guo2017OnCO}. Many uncertainty quantification methods are computationally expensive, lack compatibility with black-box models, and offer no formal guarantees. Alternatively, conformal prediction (CP) is a statistical post-processing approach that returns prediction \emph{sets} with a guarantee to cover the true label with high adjustable probability. CP only requires a held-out calibration set and offers a distribution-free model-agnostic coverage guarantee \citep{Vovk2005AlgorithmicLI,Angelopoulos2021AGI}. The model is used as a black box to compute conformity scores which capture the agreement between inputs $\vx$ and labels $y$. These prediction sets are shown to improve human decision-making both in terms of response time and accuracy \citep{Cresswell2024ConformalPS}. CP assumes exchangeability between the calibration and the test set (a relaxation of the i.i.d. assumption), making it broadly applicable to images, language models, etc. CP also applies on graph node classification \citep{Zargarbashi2023ConformalPS, Huang2023UncertaintyQO} where uncertainty quantification methods are limited. However, exchangeability, and therefore the conformal guarantee, easily breaks when the test data is noisy or subjected to adversarial perturbations.

\begin{figure}
    \centering
    \input{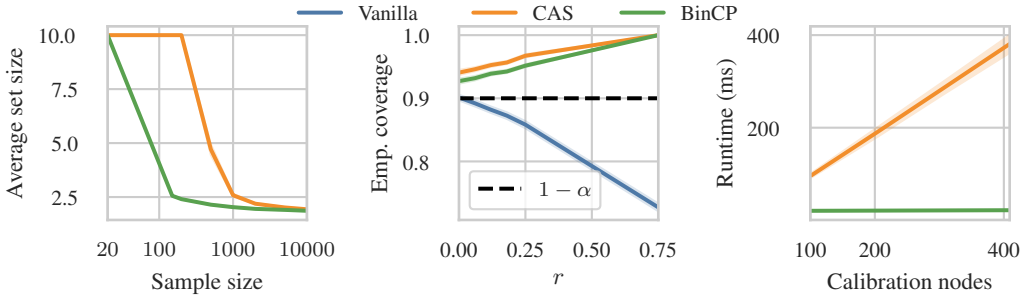}
    \caption{[Left] Average set size with different MC sample rates, 
    [Middle] empirical coverage of vanilla and robust CPs under attack,
    and [Right] runtime of robust CP as a function of calibration datapoints (after computing the MC samples which is the number of lower bound computations).}
    \label{fig:low-samples}
\end{figure}

Robust conformal prediction extends this guarantee to worst-case inputs $\pertx$ within a maximum radius around the clean point $\vx$, e.g. $\forall \pertx$ s.t. $\|\pertx - \vx\|_2 \le r$. In the evasion setting, we assume that the calibration set is clean, and test datapoints can be perturbed. Building on the rich literature of robustness certificates \citep{kumar2020certifying}, recent robust CP baselines \citep{gendler2021adversarially, zargarbashirobust, jeary2024verifiably} use a conservative score at test time that is a \emph{certified} bound on the conformity score of the clean unseen input. This maintains the guarantee even for the perturbed input since ``if CP covers $\vx$, then robust CP certifiably covers $\pertx$''. However, the average set size increases, especially if the bounds are loose.
The certified bounds can be derived through model-dependent verifiers \citep{jeary2024verifiably} or smoothing-based black-box certificates \citep{zargarbashirobust}.


For the robustness of black-box models, an established approach is to certify the confidence score through randomized smoothing \citep{kumar2020certifying}, obtaining bounds on the expected smooth score.  The tightness of these bounds depends on the information about the smooth score around the given input, e.g. the mean \citet{yan2024provably}, or the CDF \citet{zargarbashirobust}. Such methods:  \begin{enumerate*}[label=(\roman*)]
    \item assume the conformity score function has a bounded range,
    \item compute several certificates for each calibration (or test) point, and
    \item need a large number of Monte-Carlo samples to get tight confidence intervals.
\end{enumerate*} For the current SOTA method CAS \citep{zargarbashirobust}, the accounting for sample correction inflates the prediction sets significantly for sample rates below 2000 (see \subref{fig:low-samples}{left}). 
This inefficiency increases to trivially returning $\gY$ as the prediction set when we run with higher coverage rates or higher radii (see \autoref{sec:experiments}).
In contrast, we obtain robust and small prediction sets with only $\sim150$ MC samples.
Additionally, these methods require computing certified bounds for (at least) each calibration point which we further show is a wasteful computation.

\parbold{\method} We observe that smooth inference is inherently more robust.
Even without certificates, randomized methods show a slower decrease in coverage under attack (see \subref{fig:vanilla-l1-scores}{right}).  Given any score function $s(\vx, y)$ capturing conformity, \citet{zargarbashirobust} and \citet{gendler2021adversarially} define the smooth score as $\bar{s}(\vx, y) = \E_{\vepsilon \sim \gN(\boldsymbol{0}, \sigma\mI)}[s(\vx + \vepsilon, y)]$.
Instead, we perform binarization via a threshold $\tau$, i.e. $\bar{s}(\vx, y) = \E_{\vepsilon \sim \gN(\boldsymbol{0}, \sigma\mI)}[\sI[s(\vx + \vepsilon, y) \ge \tau]]=\Pr_{\vepsilon \sim \gN(\boldsymbol{0}, \sigma\mI)}[s(\vx + \vepsilon, y) \ge \tau]$. Both are valid conformity scores, and both change slowly around any $\vx$, however, our binarized CP (\method) method has several advantages.
First, we define robust CP that only computes a single certificate. In comparison, 
CAS
requires at least one certificate per calibration (or test) point.
Second, our method can effortlessly use many existing binary certificates out of the box without any additional assumptions or modifications. 
A direct consequence is that we can use de-randomization techniques  \citep{levine2021improved} that completely nullify the need for sample correction under $\ell_1$ norm.
Third, when we do need sample correction, working with
binary variables
allows us to use tighter concentration inequalities \citep{clopper1934use} (see \autoref{sec:suppl-confidence-intervals} for a detailed discussion).
     Thus, even with significantly lower MC samples, our method still produces small prediction sets (see \subref{fig:low-samples}{left}). This improvement is even more pronounced for datasets with a large number of classes (e.g. ImageNet shown in \autoref{fig:imagenet-samples-effect}).
Finally, \method does not require the score function to be bounded which is a limitation in current methods. Our code is available on the \href{https://github.com/soroushzargar/BinCP}{BinCP Github repository}.

\section{Background}
\label{sec:background}
We assume a holdout set of labeled calibration datapoints $\dcal = \{ (\vx_i, y_i) \}_{i=1}^n$ which is exchangeable with future test points $(\xtest, y_{n+1})$, both sampled from some distribution $\gD$. We have black-box access to a model from which we compute an arbitrary conformity\footnote{Conformity scores quantify agreement and are equivalent up to a sign flip to non-conformity scores.} score $s:\gX \times \gY\to\sR$, e.g. score $s(\vx, y) = \pi_y(\vx)$ where $\pi_y(\vx)$ is the predicted probability for class $y$ (other scores in \autoref{sec:algorithms}).

\parbold{Vanilla CP}  For a user-specified nominal coverage $1 - \alpha$, let 
    $q_\alpha = \quantile{\alpha}{\mathset{s(\vx_i, y_i)}_{i=1}^n \cup\{\infty\}}$ where $\quantile{\cdot}{\cdot}$ is the quantile function.
    The sets defined as 
    $\gC(\xtest) = \mathset{y: s(\xtest, y) \ge q_\alpha}$ 
    have $1 - \alpha$ guarantee to include the true label $y_{n+1}$.
Formally, $\Pr[y_{n+1} \in \gC(\xtest)] \ge 1 - \alpha$ \citep{Vovk2005AlgorithmicLI} where the probability is over $\dcal \sim \gD, \xtest \sim \gD$. 
This guarantee, and later our robust sets, are independent of the mechanics of the model and the score function -- the model’s accuracy or the quality of the score function is irrelevant. A score function that better reflects input-label agreement leads to more efficient (i.e., smaller) prediction sets. 
%
%
For noisy or adversarial inputs, the exchangeability between the test and calibration set breaks, making the coverage guarantee invalid. \subref{fig:low-samples}{middle}, and \subref{fig:vanilla-l1-scores}{right} show that an adversary (or bounded worst-case noise) can decrease the empirical coverage drastically with imperceptible perturbations on each test point. As a defense, \emph{robust} CP extends this guarantee to the worst-case bounded perturbations.

\parbold{Threat model} The adversary's goal is to decrease the empirical coverage probability by perturbing the input. Let $\gB: \gX\to  2^{\gX}$ be a ball that returns all admissible perturbed points around an input. For images a common threat model is defined by the $\ell_2$ norm: $\gB_r(\vx) = \mathset{\tilde\vx: \|\tilde\vx - \vx\|_2\le r}$ where the radius $r$ controls the perturbation magnitude. Similarly, we can use the $\ell_1$ norm. For binary data and graphs, \citet{bojchevski2020efficient} define $\gB_{r_a, r_d}(\vx) = \{\tilde\vx: \sum_{i=1}^d\sI[\tilde{\vx}_i = \vx_i - 1] \le r_d, \sum_{i=1}^d\sI[\tilde{\vx}_i = \vx_i + 1]\le r_a\}$ where the adversary is allowed to toggle at most $r_a$ zero bits, and $r_d$ one bits.

\parbold{Inverted ball $\gB^{-1}$} 
At test time we are given a (potentially) perturbed $\pertx \in \gB(\vx)$. However, to obtain robust sets, we need to reason about (the score) of the unseen clean $\vx$.
Naively, one might assume that $\vx \in \gB(\pertx)$ -- the clean point is in the ball around the perturbed point. However, this only holds in special cases such as the ball defined by the $l_2$ norm. For example, if a binary $\pertx$ was obtained by removing $r_d$ bits and adding $r_a$ bits, to able to reach the clean $\vx$ from the perturbed $\pertx$ we need to add $r_d$ bits and remove $r_a$ bits instead since $\gB_{r_a, r_d}$ unlike $\gB_r$ is not symmetric. We define the 
inverted ball $\gB^{-1}$ as the smallest ball centered at $\tilde\vx \in \gB(\vx)$ that includes the clean $\vx$. Formally, $\gB^{-1}$ should satisfy $\forall \tilde\vx \in \gB(\vx) \Rightarrow \vx \in \gB^{-1}(\tilde\vx)$. For symmetric balls like $\ell_p$-norms, $\gB^{-1} = \gB$. For the binary ball $\gB^{-1}_{r_a, r_d} = \gB_{r_d, r_a}$ we need to swap $r_a$ and $r_d$ to ensure this condition. \citet{zargarbashirobust} also discuss this subtle but important aspect without formally defining $\gB^{-1}.$


\parbold{Robust CP} Given a threat model, robust CP defines a \emph{conservative} prediction set $\bar{\gC}$  that maintains the conformal guarantee even for worst-case inputs.
Formally,
\begin{align}
\label{eq:robust-cp-guarantee}
    \Pr_{\substack{\dcal\cup\{\xtest\} \sim \gD}}[y_{n+1} \in \bar{\gC}(\xtestpert), \forall \xtestpert \in \gB(\xtest)] \ge 1 - \alpha
\end{align}
The intuition behind existing methods is as follows:
\begin{enumerate*}[label=(\roman*)]
    \item Vanilla CP covers $\xtest$ with $1 - \alpha$ probability
    \item if $y\in\gC(\xtest)$ then $y \in \bar{\gC}(\xtestpert)$
\end{enumerate*}. Thus, robust CP covers $\xtestpert$ with at least the same probability. 
Here, (ii) is guaranteed via certified lower bounds $\lowerval[s, \vx, \gB]$ or 
certified upper bounds $\upperval[s, \vx, \gB^{-1}]$.

\begin{theorem}[\textbf{Robust CP from \citet{zargarbashirobust}}]
    \label{thrm:cas-robust}
         Define $s_y(\cdot)=s(\cdot, y)$.
    With $\upperval[s_y, \tilde\vx, \gB^{-1}] \ge \max_{\vx' \in \gB^{-1}(\tilde\vx)}s(\vx', y)$, let
        $\bar{C}_\mathrm{test}(\xtestpert) = \mathset{y: \upperval[s_y, \xtestpert, \gB^{-1}] \ge q}$,
     then $\bar{\gC}_\mathrm{test}$ satisfies \autoref{eq:robust-cp-guarantee} (test-time robustness). Alternatively, with $\lowerval[s_y, \vx, \gB] \le \min_{\vx' \in \gB(\vx)}s(\vx', y)$, define ${q^{\downarrow}} = \quantile{\alpha}{\mathset{\lowerval[s_{y_i}, \vx_i, \gB]}_{i=1}^n}$. Then 
     $\bar{C}_\mathrm{cal}(\xtestpert) = \mathset{y: s(\xtestpert, y) \ge q^{\downarrow}}$ also satisfies \autoref{eq:robust-cp-guarantee} (calibration-time robustness).
\end{theorem}
In \autoref{thrm:cas-robust} test-time robustness uses $\gB^{-1}$ since it queries the clean point from the perspective of the perturbed test input. Alternatively, calibration-time robustness uses $\gB$ since the clean calibration point is given and we are finding the lower bound for the unseen test point in the test. The intuition is that the lower bound scores from the clean calibration points are exchangeable with the lower bound of the clean test input. The perturbed test input will surely have a higher score compared to this lower bound, hence it would be covered with higher probability.

We can obtain the $\lowerval, \upperval$ bounds through neural network verifiers \citet{jeary2024verifiably} or randomized smoothing \citep{cohen2019certified}. 
We focus on the latter since we get model-agnostic certificates with black-box access.
The coverage probability is theoretically proved in CP. Similarly, (adversarially) robust CP also comes with a theoretical guarantee. In both cases we can compute the empirical coverage as a sanity check. Another metric of interest in both cases is the average set size (the efficiency) of the conformal sets. 

\parbold{Randomized smoothing} 
A smoothing scheme $\xi:\gX\to\gX$ maps any point to a random nearby point. For continuous data Gaussian smoothing $\xi(\vx) = \vx + \vepsilon$ adds an isotropic Gaussian noise to the input $\vepsilon \sim \gN(\boldsymbol{0}, \sigma\mI)$. For sparse binary data \citet{bojchevski2020efficient} define sparse smoothing as $\xi(\vx) = \vx \oplus \vepsilon$ where $\oplus$ is the binary XOR, and $\vepsilon[i] \sim \mathrm{Bernoulli}(p=p_{\vx[i]})$, where $p_1$, and $p_0$ are two smoothing parameters to account for sparsity.
To simplify the notation we write $\vx + \vepsilon$ instead of $\xi(\vx)$ in the rest of the paper for both Gaussian and sparse smoothing, but our method works for any smoothing scheme beyond additive noise.
Regardless of how rapidly a score function $s(\vx, y)$ changes, the smooth score $\bar{s}(\vx, y) = \E_{\vepsilon}[s(\vx + \vepsilon, y)]$ changes slowly near $\vx$. This enables us to compute tight $\lowerval, \upperval$ bounds that depend on the smoothing strength. See \autoref{sec:robust-votecp}, \autoref{sec:randomized-smoothing-solution}, and \autoref{sec:suppl-understanding-certificates} for details. \looseness=-1

\section{Binarized Conformal Prediction (\method)}
\label{sec:method}

We define conformal sets by binarizing randomized scores. We first show that this preserves the conformal guarantee for clean data. Then in \autoref{sec:robust-votecp} we extends the guarantee to worst-case adversarial inputs. As we will see in \autoref{sec:experiments} our binarization approach has gains in terms of Monte-Carlo sampling budget, computational cost, and average set size. 

\begin{figure}[b]
    \centering
    \input{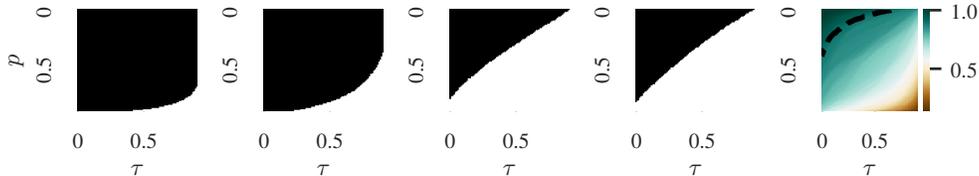}
    \caption{[Left] Function 
    $\testf(\vx_i, y_i; p, \tau)$
    for different $(p, \tau)$ pairs for four random CIFAR-10 instances. Black equals $1$ and white equals $0$. [Right] Empirical coverage for different $(p, \tau)$ pairs. Any $(p, \tau)$ pair on the dashed black line (the 0.9 contour) gives conformal sets with 90\% coverage.}
    \label{fig:accept-function}
\end{figure}

\begin{proposition}
    \label{thrm:vote-cp}
    For any two parameters $p\in (0, 1), \tau \in \sR$, given a smoothing scheme $\vx + \vepsilon$, define the boolean function $\testf[\cdot, \cdot; p, \tau]$ and the prediction set $\gC(\cdot; p, \tau)$ as
    \begin{align*}
        \testf[\vx, y; p, \tau] = \sI[\Pr_{\vepsilon} [s(\vx + \vepsilon, y) \ge  \tau] \ge p] \quad \text{and} \quad \gC(\vx; p, \tau) = \mathset{y: \testf(\vx, y; p, \tau)}
    \end{align*} 
    For any fixed $p$, let 
    \begin{align}
    \label{eq:fixed-p}
        \tau_\alpha(p) = \sup_\tau \mathset{ \tau: \sum_{i=1}^{n} \testf(\vx_i, y_i; p, \tau) \ge (1 - \alpha)\cdot(n+1)}
    \end{align}
    then the set $\gC(\xtest; p, \tau_\alpha(p))$ has $1 - \alpha$ coverage guarantee.
    Alternatively, for any fixed $\tau$, let 
    \begin{align}
    \label{eq:fixed-tau}
        p_\alpha(\tau) = \sup_p \mathset{ p: \sum_{i=1}^{n} \testf(\vx_i, y_i; p, \tau) \ge (1 - \alpha)\cdot(n+1)}
    \end{align}
    again the prediction set $\gC(\xtest; p_\alpha(\tau), \tau)$ has $1 - \alpha$ coverage guarantee.
\end{proposition}
The correctness of \autoref{thrm:vote-cp} can be directly seen by noticing that we implicitly define new scores. 

\parbold{Quantile view} Let $S_i = s(\vx_i + \vepsilon , y_i)$ be the distribution of randomized scores for $\vx_i$ and the true class $y_i$. Let $\tau_i(p) = \quantile{p}{S_i}$, we have that $\tau_\alpha(p) = \quantile{\alpha}{\{\tau_i(p)\}_{i = 1}^n}$ is a quantile of quantiles. 
Similarly, define $p_i(\tau) = \invquantile{\tau}{S_i}$ then $p_\alpha(\tau) = \quantile{\alpha}{\{p_i(\tau)\}_{i = 1}^n}$ is a quantile of inverse quantiles.
Both $\tau_i(p)$ for a fixed $p$ and $p_i(\tau)$ for a fixed $\tau$ are valid conformity scores for the instance $\vx_i$, since exchangeability is trivially preserved.
Therefore, $\tau_\alpha(p)$ and $p_\alpha(\tau)$ are just the standard quantile thresholds from CP on some new score functions. This directly gives the $1 - \alpha$ coverage guarantee.
This view via the implicit scores is helpful for intuition, but we keep the original formulation since it is more directly amenable to certification as we show in \autoref{sec:robust-votecp}.
We provide an additional formal proof of \autoref{thrm:vote-cp} via conformal risk control \citep{angelopoulos2022conformal} in \autoref{sec:proofs}.

Using either variant from \autoref{thrm:vote-cp} let $(p_\alpha, \tau_\alpha)$ equal $(p, \tau_\alpha(p))$ or $(p_\alpha(\tau), \tau)$ as the final pair of parameters. For test points $\xtest$ we accept labels whose smooth score distribution has at least $p_\alpha$ proportion above the threshold $\tau_\alpha$, i.e. $\testf(\xtest, y; p_\alpha, \tau_\alpha ) = 1$. 
The term ``binarization'' refers to mapping each score sample above $\tau$ to $1$ and all others $0$. For distributions with a strictly increasing and continuous CDF (e.g. isotropic Gaussian smoothing) both variants are equivalent.

\begin{lemma}
    \label{thrm:views-equal}
    Given distributions $\{S_i\}_{i=1}^n$ with strictly increasing and continuous CDFs, let $\tau_\alpha(p)$ be obtained from \autoref{eq:fixed-p} with fixed $p$ and $p_\alpha(\cdot)$ be as defined in \autoref{eq:fixed-tau}. We have $p_\alpha(\tau_\alpha(p)) = p$.
\end{lemma}

We defer all proofs to \autoref{sec:proofs}.  
For fixed $p$, \autoref{thrm:vote-cp} yields $(p, \tau_\alpha(p))$. 
Fixing $\tau=\tau_\alpha(p)$ we get  sets with $(p_\alpha(\tau), \tau)=(p_\alpha(\tau_\alpha(p)), \tau_\alpha(p))$
which also equals
$(p, \tau_\alpha(p))$ from \autoref{thrm:views-equal}.
\autoref{fig:accept-function} shows the $\testf(\vx, y; p, \tau)$ function for several examples. This function is non-increasing in both parameters $p$ and $\tau$. 
In general, any arbitrary assignment of $p$, and $\tau$, results in some expected coverage -- $\testf(\cdot, \cdot, p, \tau)$ equals to 1 for some number of $(\vx_i, y_i)$s (\subref{fig:accept-function}{right}). Pairs $(p_\alpha, \tau_\alpha)$ obtained from \autoref{thrm:vote-cp} are placed on the $1 - \alpha$ contour of this expectation. 
The empirical coverage is close to this expectation due to exchangeability \citep{berti1997glivenko}.

\textbf{Remarks.}
The scores $\tau_i(p)$ (and similarly $p_i(\tau)$ remain exchangeable whether the quantile over the smoothing distribution is computed exactly or estimated from any number of  Monte-Carlo samples. That is, \autoref{thrm:vote-cp} holds regardless. However, when need to be more careful when we consider the certified upper and lower bounds.
In  \autoref{sec:robust-votecp} we first derive robust conservative sets that maintain worst-case coverage, assuming that we can compute probabilities and expectations exactly. Since this is not always possible, 
in \autoref{sec:monte-carlo} we provide the appropriate sample correction that still preserves the robustness guarantee when using Monte-Carlo samples. We also discuss a de-randomized approach that does not need sample correction.

\section{Robust \method}
\label{sec:robust-votecp}

From \autoref{thrm:vote-cp} (either variant) we compute a pair $(p_\alpha, \tau_\alpha)$.
Following \autoref{thrm:vote-cp}, for clean $\xtest$, we have $\Pr[s(\xtest + \vepsilon, y_{n+1}) \ge \tau_\alpha] \ge p_\alpha$ with probability $1 - \alpha$. We will exploit this property. Define $f_y(\vx) = \sI[s(\vx, y) \ge \tau_\alpha]$, we have $\bar{f}_y(\vx) = \E_{\vepsilon}[\sI[s(\vx + \vepsilon, y) \ge \tau_\alpha]] = \Pr_{\vepsilon}[s(\vx+\vepsilon, y) \ge \tau_\alpha]$.

\parbold{Conventional robust CP} One way to attain robust prediction sets is to apply the same recipe as \citet{zargarbashirobust} (CAS) by finding upper or lower bounds on the new score function. CAS uses the smooth score $\bar{s}_y(\vx) = \E_{\vepsilon}[s(\vx + \vepsilon, y)]$. Instead, we can bound $\bar{f}_y(\vx)$ which is a smooth binary classifier. 
Note that as discussed in \autoref{sec:method} (quantile of inverse quantiles), $\bar{f}_y(\vx)$ is a conformity score function itself. 
Therefore, following \autoref{thrm:cas-robust}, the test-time, and calibration-time robust prediction sets are \begin{equation}
    \bar\gC_\mathrm{test}(\xtestpert) = \{y: \upperval[\bar{f}_y, \xtestpert, \gB^{-1}] \ge p_\alpha\}, \quad \bar\gC_\mathrm{cal}(\xtestpert) = \{y: \bar{f}_y(\xtestpert)  \ge q^{\downarrow}\}
\end{equation}
where $q^{\downarrow} = \quantile{\alpha}{\{\lowerval[\bar{f}_{y_i}, \vx_i, \gB]\}_{i=1}^n}$. In short, we replace the clean
$\bar{f}_{y_{n+1}}(\xtest)$
with either its certified upper $\upperval$ or lower $\lowerval$ bound. We elaborate on this approach before improving it.

\parbold{Computing $\lowerval$ and $\upperval$}
Computing exact worst-case bounds on $\bar{f}$ ($\bar{f}_y$ for all $y$) is intractable and requires white-box access to the score function and therefore the model. Following established techniques in the randomized smoothing literature \citep{lee2019tight}
we relax the problem.
Formally,\begin{equation}
    \label{eq:upperbound-binary}
    \lowerval[\bar{f}, \vx, \gB] = \min_{\substack{\tilde\vx \in \gB(\vx) \\ h \in \gH}} \Pr_{\vepsilon}[h(\tilde\vx + \vepsilon)] \quad \text{s.t.} \quad \Pr_{\vepsilon}[h(\vx + \vepsilon)]  = \Pr_{\vepsilon}[f(\vx + \vepsilon)] =\bar{f}(\vx)
\end{equation} 
where $\gH$ is the set of all measurable functions $h$. Since $f \in \gH$ we have
$\lowerval[\bar{f}, \vx, \gB] \leq \bar{f}(\pertx)$ for all $\pertx \in \gB(\vx)$.
The upper bound $\upperval[\bar{f}, \vx, \gB^{-1}]$ is the solution to a similar \emph{maximization} problem.

\parbold{Closed form} For $\ell_2$ ball with Gaussian smoothing, \autoref{eq:upperbound-binary} has a closed form solution $\Phi_\sigma(\Phi_\sigma^{-1}(\bar f_y(\vx))-r)$ where $\Phi_\sigma$ is the CDF of the normal distribution $\gN(\boldsymbol{0}, \sigma\mI)$\citep{cohen2019certified, kumar2020certifying}. The upper bound is similarly computed by changing the sign of $r$. \citet{yang2020randomized} show the same closed-form applies solution for the $\ell_1$ ball, and additionally, discuss other perturbation balls and smoothing schemes most of which are applicable. For sparse smoothing we can compute the bounds with a simple algorithm with $O(r_a+r_d)$ runtime \citep{bojchevski2020efficient}, which we discuss in \autoref{sec:proofs}. For $\ell_1$ ball and uniform smoothing the lower bound equals $\bar f_y(\vx)-1/(2\lambda)$ where $\vepsilon \sim \gU[0, 2\lambda]^d$ \citep{levine2021improved}. This bound can also be de-randomized (see \autoref{sec:monte-carlo}).

\parbold{Single Binary Certificate} From the closed-form solutions we see that the bounds are independent of the definition of $f$,
and the test point $\vx$; i.e. their output is a function of the scalar $p:=\bar{f}_y(\vx)$. 
We defer the discussion for why this holds to \autoref{sec:randomized-smoothing-solution}, and \autoref{sec:suppl-understanding-certificates}; in short the solution for any $\vx$ can be obtained from alternative canonical points $\vu$, and $\tilde{\vu}$.
Therefore, we write $\lowerval[p, \gB]= \lowerval[\bar{f}_y, \vx, \gB]$ to show that $\lowerval$ depends only on $p$ and $\gB$, and the same for $\upperval$. 
We also notice that in common smoothing schemes and perturbation balls, it holds that $\lowerval[\upperval[p, \gB^{-1}], \gB]= p$ which allows us to reduce both calibration-time and test-time robustness to solving a single binary certificate. We formalize this in \autoref{thrm:robust-reduced}. \looseness=-1

\begin{lemma}
\label{thrm:robust-reduced}
    If  $\lowerval[\upperval[p, \gB^{-1}], \gB]= p$ for all $p$, then $\bar{\gC}_\mathrm{test}(\xtestpert) = \bar{\gC}_\mathrm{cal}(\xtestpert) = \bar{\gC}_\mathrm{bin}(\xtestpert)$ where $\bar{\gC}_\mathrm{bin} (\xtestpert) = \{y: \testf(\xtestpert, y; \lowerval[p_\alpha, \gB], \tau_\alpha) \} = \{ y: \Pr_{\vepsilon}[s(\xtest + \vepsilon, y_{n+1}) \ge \tau_\alpha] \ge   \lowerval[p_\alpha, \gB] \}$. 
\end{lemma}

\begin{figure}[t]
    \centering
    \input{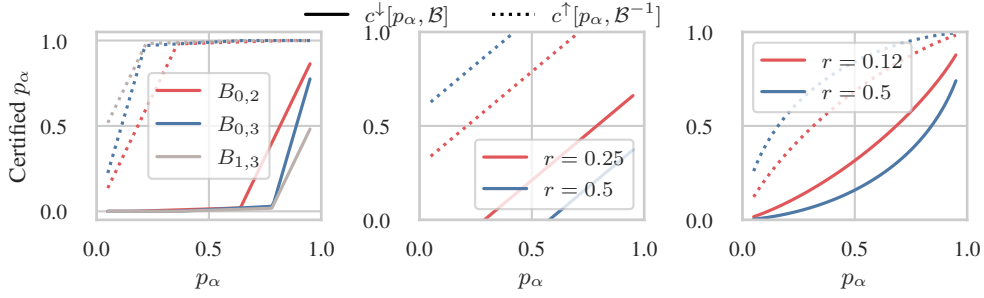}
    \caption{[From left to right] Certified bounds for sparse smoothing, $\ell_1$ ball with de-randomized uniform smoothing \citep{levine2021improved}, and $\ell_2$ (same as $\ell_1$) ball with Gaussian smoothing.\looseness=-1}
    \label{fig:average-size-all}
    \label{fig:p-upper-lower}
\end{figure}

To see why, 
let $\tilde{p}_{n+1} = \bar{f}_{y_{n+1}}(\xtestpert)$. The test-time robust coverage requires $\upperval[\tilde{p}_{n+1}, \gB^{-1}] \ge p_\alpha$. Since both $\lowerval$, and $\upperval$ are non-decreasing w.r.t. $p$, we have $\lowerval[\upperval[\tilde{p}_{n+1}, \gB^{-1}], \gB] \ge \lowerval[p_\alpha, \gB]$. We have the equivalent condition $\tilde{p}_{n+1} \ge \lowerval[p_\alpha, \gB]$. This implies that we only need to compute a single certificate $\lowerval[p_\alpha, \gB]$ once with the single $p_\alpha$ value given by \autoref{thrm:vote-cp}. This also allows us to seamlessly integrate other existing binary certificates in a plug and play manner. 
In contrast, with \autoref{thrm:cas-robust} for $\bar{\gC}_\mathrm{test}$ or $\bar{\gC}_\mathrm{cal}$ we need at least one certificate per test (or calibration) point. Notably, these prediction set are identical to our cheaper $\bar\gC_\mathrm{bin}$.
For illustration, \autoref{fig:p-upper-lower} shows the certified lower and upper bounds for various $p_\alpha$ values and various smoothing schemes.


Intuitively $\lowerval[\upperval[p, \gB^{-1}], \gB]= p$ holds due to symmetry of the smoothing scheme w.r.t. $\gB$, and $\gB^{-1}$ and is satisfied by most smoothing schemes. In \autoref{thrm:gaussian-sparse-symmetric} we prove that Gaussian, uniform, and sparse smoothing all have this property. 

\begin{lemma}
    \label{thrm:gaussian-sparse-symmetric}
    For Gaussian, and uniform smoothing under $\ell_1$, and $\ell_2$ balls $\gB_r = \gB^{-1}_r$.  For sparse smoothing and $\gB_{r_a, r_d}$ we have $\gB^{-1}_{r_a, r_d} = \gB_{r_d, r_a}$. In all three cases we have $\lowerval[\upperval[p, \gB^{-1}], \gB]= p$.\looseness=-1
\end{lemma}

To summarize, for robust \method, we first compute conformal thresholds $(p_\alpha, \tau_\alpha)$ from \autoref{thrm:vote-cp}. Then for a perturbation ball $\gB$ that satisfies $\lowerval[\upperval[p, \gB^{-1}], \gB]= p$, we compute $\lowerval[p_\alpha, \gB]$ and compute the prediction sets with $(\lowerval[p_\alpha, \gB], \tau_\alpha)$ instead. The resulting sets have $1- \alpha$ robust coverage. 

\begin{corollary}
\label{thrm:bincp-final}
    With $(p_\alpha, \tau_\alpha)$  from \autoref{thrm:vote-cp} on a calibration set $\dcal$, let $\xtest$ be exchangeable with $\dcal$ and $\xtestpert \in \gB(\xtest)$. If for the smoothing scheme $\xi$ and the threat model $\gB$ and for all $p$ we have $\lowerval[\upperval[p, \gB^{-1}], \gB]= p$, then the set $\bar{\gC}_\mathrm{bin} (\xtestpert) = 
    \{ y: \Pr[s(\xtestpert + \vepsilon, y_{n+1}) \ge \tau_\alpha] \ge   \lowerval[p_\alpha, \gB] \}
    $ has $1 - \alpha$ coverage (the pseudocode is in \autoref{sec:algorithms}).
\end{corollary}

\section{Robust \method with Finite Samples}
\label{sec:monte-carlo}

The certificate in \autoref{thrm:bincp-final} relies on exact probabilities $ \Pr[s(\xtest + \vepsilon, y_{n+1}) \ge \tau_\alpha]$ which is often intractable to compute. Instead, we can either apply de-randomization techniques or estimate high-confidence bounds of these probabilities. We first describe the latter approach.
For each calibration point $(\vx_i, y_i)$ we compute $q_i = \frac{1}{m}\sum_{i = 1}^{m}\sI[s(\vx_{i} + \vepsilon, y_i) \ge \tau_\alpha]$ where $m$ is the number of Monte-Carlo (MC) samples.
For each label of the (potentially perturbed) test point we compute $\tilde{q}_{n+1, y} = \frac{1}{m}\sum_{i = 1}^{m}\sI[s(\xtestpert + \vepsilon, y) \ge \tau_\alpha]$. 
We use the Clopper-Pearson confidence interval \citep{clopper1934use} to bound the exact probabilities via the MC estimates.
To ensure the sets are conservative we compute a lower bound 
for calibration points and an upper bound for test points. 
Collectively, all bounds are valid with adjustable $1 - \eta$ probability. To account for this, we set the nominal coverage level to $1 - \alpha + \eta$ such that we have $1 - \alpha$ coverage in total. Similar to \citet{zargarbashirobust}, we compute each bound with $1 - \eta / (|\dcal + k|)$ probability where $k$ is the number of classes. 
Let $p_i = \Pr[s(\vx_i + \vepsilon, y_i) \ge \tau_\alpha]$ for $i \in \{1, \dots, n+1\}$ be the exact probabilities. The final sample-corrected robust predictions sets are given in \autoref{thrm:MC-tau-fixed}.
\begin{proposition}
    \label{thrm:MC-tau-fixed}
    Let $q_{i}^\downarrow \le p_i$ hold with $1 - \eta/(\dcal + k)$ for each calibration point $i \in \{1, \dots, n\}$ where $k$ is the number of target classes. For a given test point $\xtestpert$ let $\tilde{q}_{n+1, y}^\uparrow \ge \tilde{p}_{n+1, y}$ with $1 - \eta/(\dcal + k)$ where $\tilde{p}_{n+1, y} = \Pr[s(\xtestpert + \vepsilon, y) \ge \tau_\alpha]$. With $p_{\alpha}^\downarrow = \quantile{\alpha - \eta}{\{q_{i}^\downarrow\}_{i=1}^n}$, we set the robust conformal threshold pair as $(\lowerval[p_{\alpha}^\downarrow, \gB], \tau_\alpha)$. Then the prediction set defined as $\bar\gC_{+}(\xtestpert; \lowerval[p_{\alpha}^\downarrow, \gB], \tau_\alpha) = \{y: \tilde q_{n+1, y}^\uparrow \ge \lowerval[p_{\alpha}^\downarrow, \gB]\}$ has $1 - \alpha$ coverage probability.
\end{proposition}
Such sample correction is a crucial step for smoothing-based robust CP, since the robustness certificate is probabilistic. The failure of the certificate depends to the failure of the confidence intervals. In contrast, for deterministic and de-randomized certificates such as DSSN \citep{levine2021improved}, we do not need sample correction since we can exactly compute $p_i$  and $p_\alpha = \quantile{\alpha}{\{p_i\}_{i=1}^n}$.
Note, vanilla (non-robust) \method does not need sample correction to maintain the guarantee (see \autoref{sec:method}). 

\section{Experiments}
\label{sec:experiments}

We show that: \begin{enumerate*}[label=(\roman*)]
    \item We can return guaranteed and small sets for both image classification and node classification, with a significantly lower number of Monte Carlo samples. 
    \item Our sets are computationally efficient.
    \item There is an inherent robustness in randomized methods. 
    \item We can also use de-randomized smoothing-based certificates that do not require finite sample correction. 
\end{enumerate*} 

\parbold{Setup} We evaluate our method on two image datasets: CIFAR-10 \citep{Krizhevsky2009LearningML} and ImageNet \citep{Deng2009ImageNetAL}, and for node-classification (graph) dataset we use Cora-ML \cite{McCallum2004AutomatingTC}. For the CIFAR-10 dataset we use \texttt{ResNet-110} and for the ImageNet dataset we use ResNet-50 pretrained models with noisy data augmentation from \citet{cohen2019certified}. For the graph classification task we similarly train a GCN model \cite{Kipf2017SemiSupervisedCW} on CoraML with noise augmentation. The GCN is trained with 20 nodes per class with stratified sampling as the training set (and similarly sampled validation set). 
The size of the calibration set is between 100 and 250 (sparsely labeled setting) unless specified explicitly. Our reported results on conformal prediction performance are averaged over 100 runs with different calibration set samples. We calibrated \method with a $p = 0.6$ fixed value, however small changes in $p$ do not influence the result. For the graph dataset we calibrated \method with $p=0.9$. Intuitively as $\lowerval[p, \gB]$ has a sharp decay for sparse smoothing (see \subref{fig:conf-radius}{left}), we set $p$ to a number such that $\lowerval[p, \gB]$ still remains high.

\begin{figure}[t]
    \centering
    \input{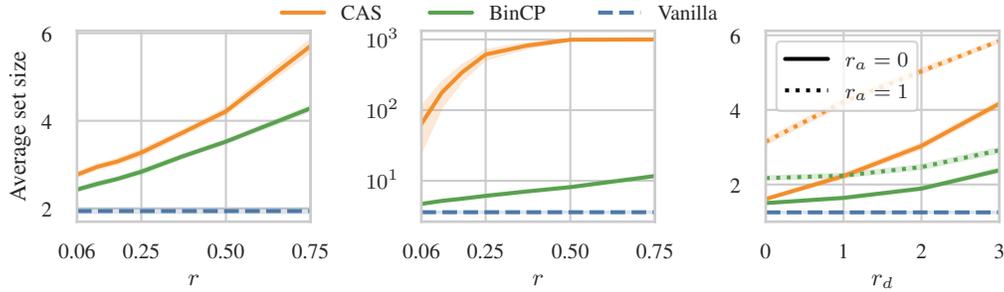}
    \caption{[Left to right] Average prediction set size of robust CP for CIFAR-10, and ImageNet with Gaussian smoothing ($\sigma=0.5$), and CoraML with sparse smoothing. All results are for 2000 Monte-Carlo samples. We set $1 - \alpha = 0.85$ for ImageNet,
    and $1 - \alpha = 0.9$ for CIFAR-10 and CoraML.}
    \label{fig:average-size-all}
\end{figure}
\begin{figure}[ht!]
    \centering
    \input{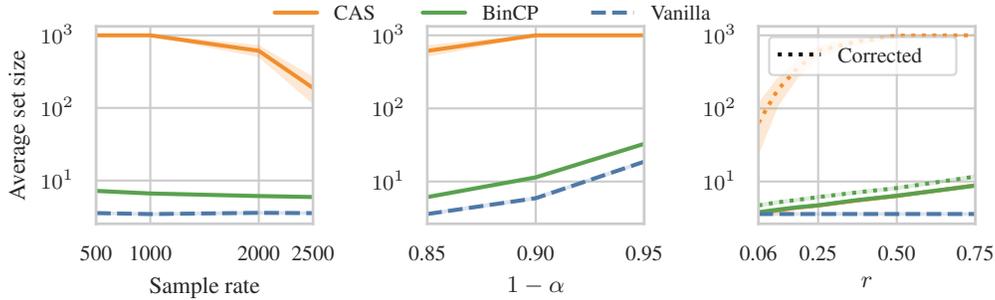}
    \caption{On ImageNet dataset, [left] average set size for $1 - \alpha=0.85$ with various MC sampling budgets. [Middle] Set size across various levels of $1 - \alpha$ for $2\times10^3$ samples. [Right] Set size without sample correction (asymptotically valid assumption). The sample-corrected variants are shown with a dotted line. In all plots $y$-axis is log-scaled.}
    \label{fig:imagenet-samples-effect}
\end{figure}

We conducted our experiment using three different smoothing schemes. \begin{enumerate*}[label=(\roman*)]
    \item Smoothing with isotropic Gaussian noise, $\sigma=0.12$, $0.25$, and $0.15$. Our reported results for \method are valid for both $\ell_1$, and $\ell_2$ perturbation balls.   
    \item De-randomized smoothing with splitting noise (DSSN) from \citet{levine2021improved} from which we attain $\ell_1$ robustness. We examine two smoothing levels $\lambda=0.25 / \sqrt{3}$, and $0.5 / \sqrt{3}$.
    \item Sparse smoothing from \citet{bojchevski2020efficient} with $p_+=0.01$, and $p_-=0.6$ on node attributes. We report robustness across $r_a \in \{0, 1\}$, and $r_d \in \{0, 1, 2, 3\}$.
\end{enumerate*}
We compare our the result from \method to the SOTA method CAS \citep{zargarbashirobust}. Previously it was shown that CAS significantly outperforms RSCP \citep{gendler2021adversarially} both with and without  finite sample correction. In \autoref{sec:related-works} we discuss the other related works in detail. In the standard setup, we estimate the statistics (mean and CDF, or Bernoulli parameters) with $2\times 10^3$ Monte-Carlo samples, and we set $1 - \alpha = 0.9$. This setup is picked in favor of the baseline since by increasing the nominal coverage or decreasing the sample size \method outperforms the baseline with an even higher margin.  Throughout the paper we report different nominal coverages and MC sampling budgets. 

\begin{figure}[b]
    \centering
    \input{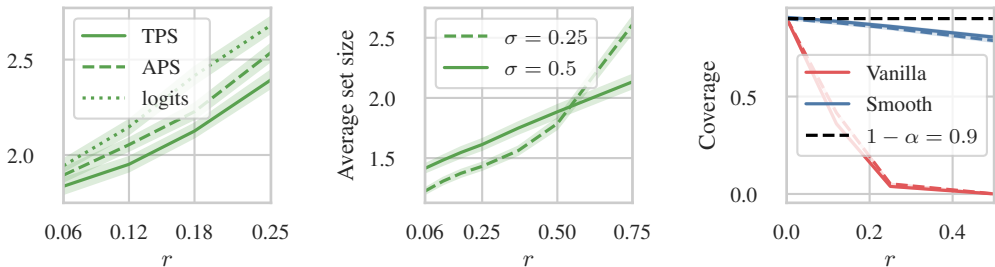}
    \caption{[Left] Performance of \method on different score functions under Gaussian smoothing with $\sigma=0.25$. [Middle] Set size of \method with $\ell_1$ robustness and derandomized DSSN smoothing ($\sigma=\lambda/\sqrt{3}$). [Right] Vanilla non-smooth and smooth ($\sigma=0.25$) prediction (solid and dashed colored lines show TPS and APS score function) under attack. All results are on CIFAR-10.}
    \label{fig:vanilla-l1-scores}
\end{figure} 

\begin{figure}
    \centering
    \input{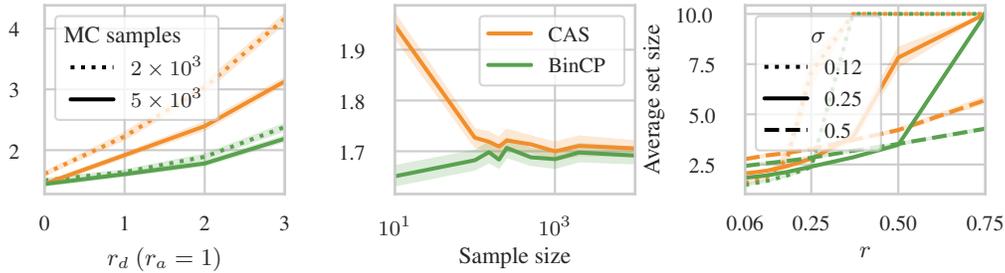}
    \caption{Comparison between \method and CAS for [Left] the effect of higher MC sample budget in CoraML dataset (sparse smoothing), [Middle] effect of low samples without finite samples correction for CIFAR-10 dataset ($\sigma=0.25$), and [Right] various smoothing strengths $\sigma$.}
    \label{fig:ablations}
\end{figure}

\parbold{Smaller set size} \autoref{fig:average-size-all} shows that for all datasets, and both smoothing schemes (isotropic Gaussian and sparse smoothing), \method produces smaller prediction sets compared to CAS. Our prediction sets computed with Gaussian noise are robust to both $\ell_1$, and $\ell_2$ perturbation balls with the same radius \citep{yang2020randomized}.
In \autoref{fig:imagenet-samples-effect}, compare to CAS on ImageNet dataset over various sample rates, coverages, and radii. Since CAS with sample correction returns trivial sets for $1 - \alpha=0.9$ (see \subref{fig:imagenet-samples-effect}{middle}), in \subref{fig:imagenet-samples-effect}{left} we compare the results for $1 - \alpha=0.85$. 
%
By increasing the Monte Carlo sample rate, the average set sizes of CAS and BinCP become closer
--
\subref{fig:low-samples}{left} for CIFAR-10, \subref{fig:imagenet-samples-effect}{left} for ImageNet, and \subref{fig:ablations}{left} for CoraML depict the impact of higher sampling budget. Intuitively as the number of classes increase (e.g. ImageNet dataset) the union bound over classes result in larger prediction sets (looser confidence intervals). 
In \autoref{sec:additional-expr} (\autoref{fig:low-r}) we show that \method is also consistently better for smaller radii (a setup similar to verifier based robust CP \citep{jeary2024verifiably}). For the same reported set size, we attain robust CP of significantly larger radii. \looseness=-1 

Note that \subref{fig:average-size-all}{right}, and \subref{fig:ablations}{left} show the performance on the \emph{transductive} node classification 
setting where perfect robustness is achievable for free through memorization
(see discussion by \citet{gosch2023adversarial}).
Nonetheless, comparing \method to CAS is still meaningful.
Constructing robust conformal sets for GNNs in a realistic (inductive) setup is more challenging as discussed in \autoref{sec:suppl-gnn-setup}.

\parbold{Exact $\ell_1$ robustness} 
Using the de-randomized DSSN certificate for $\ell_1$ perturbation ball \citep{levine2021improved} we derive the first smoothing-based de-randomized robust CP.
As shown in \subref{fig:vanilla-l1-scores}{middle} the de-randomized robust \method with uniform noise results in a significantly smaller set size across all radii compared to Gaussian noise (\subref{fig:average-size-all}{left}). Notably due to exactness of the computed statistics, for randomized DSSN-based certificate we bypass the finite samples correction.
%

\parbold{Ignoring sample correction} 
While unrealistic in practice, \citet{gendler2021adversarially} report results without applying finite sample correction.  \citet{zargarbashirobust} maintain small set sizes (with large MC sample rate) for CIFAR-10. However, for ImageNet and CoraML they only report the results without correction, and therefore with an ``asymptomatically valid'' coverage guarantee -- valid when sample rate approaches infinity.
In \subref{fig:imagenet-samples-effect}{left} applying sample correction to CAS on datasets like ImageNet, inflates the prediction sets up to $\gY$, likely due to union bound on a large number of classes. 
\subref{fig:imagenet-samples-effect}{right} shows that on ImageNet, both methods show similar prediction sets for asymptotically valid setup. Notably  \method with sample correction is not far from the non-corrected setup, while CAS shows a large gap. Similarly \subref{fig:ablations}{left} shows that for sparse smoothing, increasing sample rate helps CAS considerably while its effect on \method is almost negligible. In \autoref{sec:suppl-confidence-intervals} we explore the intuition behind how binarization can mitigate the impact of finite sample correction with fewer samples. \looseness=-1

\parbold{Number of samples}
The upperbound in CAS is obtained through a two step process. First given the corrected CDF, we compute the worst case (adversarial) CDF. Then using upper bounded (or lower bounded) CDF, we apply the Anderson bound to obtain a bound on the mean from the CDF \citep{zargarbashirobust}. Increasing the number of bins increases the computation slightly but produces tighter bounds. To observe this effect, without sample correction, we decrease the number of samples to a very low number ($\sim 10$, however unrealistic) and in \subref{fig:ablations}{middle} we see that set size in CAS slightly increases even without accounting for finite sample estimation.

\parbold{Effect of $\sigma$, and score function}
The strength of Gaussian smoothing is controlled by $\sigma$ in $\xi(\vx) = \vx + \vepsilon, \vepsilon\sim \gN(\boldsymbol{0}, \sigma^2\mI)$. 
In \subref{fig:ablations}{right} we observe a trade-off in choosing $\sigma$ for both methods. Higher smoothing intensity results in larger set sizes in the beginning, but by increasing the robustness radius the set size increases slowly.
Still in all cases \method outperforms CAS. It is best practice to compute smooth prediction probabilities using a model trained with similar noise augmentation. We reported this result in \autoref{sec:additional-expr} (\autoref{tab:all-sigmas}). Interestingly, when training and inference $\sigma$ does not match, \method shows good results while CAS is more sensitive.
%

We mainly focus on TPS score function, in addition \subref{fig:vanilla-l1-scores}{left} compares APS, and logit as score functions. Here across all radii TPS is more efficient. We further report results for APS in \autoref{sec:additional-expr}. Since logits are unbounded, CAS is not applicable to it.

\parbold{Benefits of smoothing} The guarantee of robust CP breaks for adversarial (or noisy) inputs. In \subref{fig:vanilla-l1-scores}{right} we compare vanilla prediction and smooth prediction sets under adversarial attack. Notably, smooth models even without a conservative certificate show an inherent robustness. As illustrated, the non-smooth model quickly breaks to near 0 coverage guarantee for very small $r$.
Relatedly, recent verifier-based robust CP \citep{jeary2024verifiably} report comparably larger prediction sets even for one order of magnitude smaller radius (compared to the certified radii by \method). This intuitively suggests that for robust CP it seems that randomization is inherently beneficial.

\parbold{Limitations} 
    While \method reduce the required MC sample rate significantly (e.g. from 2000 to 150 on CIFAR-10), still this number of inferences is computationally intensive.
    The robustness in the input space $\gX$ is not yet linked to robustness w.r.t. distribution shift.
    Although \method applies on sparse smoothing, a realistic threat model for graphs (inductive GNNs) is still not addressed.

\section{Related Work}
\label{sec:related-works}

\parbold{Robust CP via smoothing} \citet{gendler2021adversarially} introduced the problem and defined a baseline robust CP method, RSCP (randomly smoothed conformal prediction), which applies \autoref{thrm:cas-robust} in combination with the mean-constrained upper bound for $\ell_2$ perturbations and Gaussian smoothing. This upper bound has a closed form solution: $\overline{s}(\tilde\vx, y) = \Phi(\Phi^{-1}(p) + r)$ where $p = \E[s(\vx + \vepsilon, y)]$. Originally, RSCP did not account for finite sample correction making its coverage guarantee only asymptomatically valid. \citet{yan2024provably} show that correcting for finite samples in RSCP leads to trivial prediction sets $\bar\gC(\vx) = \gY$. As a remedy, they define a new score function based on temperature scaling which in combination with conformal training \citep{stutz2021learning} improves the average set size. 
So far both methods use test-time robustness.
In \autoref{sec:additional-expr}, we show that \method outperforms RSCP+.

In contrast \citet{zargarbashirobust} utilizes the CDF structure of the score and instead apply the tighter CDF-based bound defining CDF aware sets (CAS). In combination with calibration-time robustness, they show that only $|\dcal|$ certificate bounds should be computed to maintain a robust coverage guarantee as in \autoref{eq:robust-cp-guarantee}. In addition to a gain in computational efficiency, they show that in the calibration-time robustness, the error correction budget can be used more efficiently. On CIFAR-10 they return a relatively small conformal set size. In all aforementioned methods, a large MC sampling budget (e.g. $10^4$ samples) is assumed which is challenging for real-time applications. This issue is exacerbated for datasets like ImageNet where the large number of classes amplifies the effect of multiple testing corrections. 

\parbold{Robust CP via verifiers} Outside the scope of randomized smoothing \citet{jeary2024verifiably} use neural network verification to compute upper (or lower) bounds. 
This requires white-box access to the model weights, while our proposed method works for any black-box model and randomized or exact smoothing-based certificate. Interestingly, in \citep{jeary2024verifiably} (Table 1) the empirical evaluation is for $r=0.02$ which is smaller than the minimum radius we reported. For completeness, we evaluated \method on very small radii in \autoref{sec:additional-expr} (\subref{fig:low-r}{left}), and for the same $r$ our sets are $2\times$ smaller. As discussed in \autoref{sec:experiments} (\autoref{fig:vanilla-l1-scores}) in general, smooth prediction, even without accounting for the adversary, shows to have an inherent robustness. 

\parbold{Other robustness results}  Alternatively \citet{Ghosh2023ProbabilisticallyRC} introduce probabilistic robust coverage which intuitively accounts for \emph{average} adversarial inputs. This is in contrast with our core assumption of worst-case adversarial inputs. In other words, instead of $1 - \alpha$ coverage for any point within the perturbation ball around $\xtest$, ``probabilistically robust coverage'' guarantees that the probability to cover the true label remains above $1 - \alpha$ on average over all $\tilde\vx \in \gB(\vx)$, while we consider the worst-case $\tilde\vx$. Their ``quantile of quantiles'' method looks superficially similar to \method as they also compute $n+k+1$ quantiles. However there are two notable differences.
    Their first order of quantiles (on true calibration scores and the score for each class of the test point) is over random draws from the perturbation set. \method computes the first order quantiles ($\tau_i(p)$ in fixed $\tau$ setup) over the smooth score distribution. 
    Their conservative quantile index is based on a user-specified hyperparameter that accounts for conservativeness while \method finds the certified probability $\lowerval[p_\alpha, \gB]$ for the worst case adversarial example. \method guarantees that any $\tilde\vx\in\gB(\vx)$ is covered if $\vx$ is covered.
Furthermore, there are other works addressing distribution or covariate shift in general beyond the score of worst-case noise robustness \citep{Barber2022ConformalPB, tibshirani2019conformal}.\looseness=-1

\section{Conclusion}
We introduce \method, a robust conformal prediction method based on randomized smoothing that produces small prediction sets with a few Monte-Carlo samples. The key insight is that we binarize the distribution of smooth scores, by a threshold (or thresholds) that maintains the coverage guarantee. We show that both calibration and test-time robustness approaches (discussed in \autoref{sec:background}) are equivalent to computing a single binary certificate. This directly enables us to use any certificate that returns a certified lower-bound probability right out of the box; including de-randomized certificate for $\ell_1$ norm. The binarization enables us to use tighter Clopper-Pearson confidence intervals. This leads directly to faster computation of prediction sets with a significantly lower Monte Carlo sample-rate (compared to the SOTA), and therefore less forward passes per input. 
Interestingly, we show that even without accounting for an adversarial setup, CP with smooth score shows more robustness to adversarial examples in comparison with the conventional vanilla CP. 

\section*{Acknowledgment}
We thank Giuliana Thomanek and Jimin Cao for their feedbacks on our initial draft.

\section*{Ethics statement}
In this paper, we study the robustness of conformal prediction. The main focus of our work is to increase the reliability of conformal prediction in presence of noise or adversarial perturbations. Therefore, we don't see any particular ethical concern to mention about this study.

\section*{Reproducibility statement}
To ensure the reproducibility of our results, we have provided the algorithm is in \autoref{sec:algorithms}, and the implementations are available at \href{https://github.com/soroushzargar/BinCP}{the BinCP Github repository}. The models we used are also pre-trained and all accessible from the cited works. We specified the setup including parameter selections in \autoref{sec:experiments}.

\bibliographystyle{iclr2025_conference}
\bibliography{iclr2025_conference}

\begin{thebibliography}{31}
\providecommand{\natexlab}[1]{#1}
\providecommand{\url}[1]{\texttt{#1}}
\expandafter\ifx\csname urlstyle\endcsname\relax
  \providecommand{\doi}[1]{doi: #1}\else
  \providecommand{\doi}{doi: \begingroup \urlstyle{rm}\Url}\fi

\bibitem[Angelopoulos et~al.(2022)Angelopoulos, Bates, Fisch, Lei, and Schuster]{angelopoulos2022conformal}
Anastasios~N Angelopoulos, Stephen Bates, Adam Fisch, Lihua Lei, and Tal Schuster.
\newblock Conformal risk control.
\newblock \emph{arXiv preprint arXiv:2208.02814}, 2022.

\bibitem[Angelopoulos \& Bates(2021)Angelopoulos and Bates]{Angelopoulos2021AGI}
Anastasios~Nikolas Angelopoulos and Stephen Bates.
\newblock A gentle introduction to conformal prediction and distribution-free uncertainty quantification.
\newblock \emph{ArXiv}, abs/2107.07511, 2021.
\newblock URL \url{https://api.semanticscholar.org/CorpusID:235899036}.

\bibitem[Barber et~al.(2022)Barber, Cand{\`e}s, Ramdas, and Tibshirani]{Barber2022ConformalPB}
Rina~Foygel Barber, Emmanuel~J. Cand{\`e}s, Aaditya Ramdas, and Ryan~J. Tibshirani.
\newblock Conformal prediction beyond exchangeability.
\newblock \emph{The Annals of Statistics}, 2022.
\newblock URL \url{https://api.semanticscholar.org/CorpusID:247158820}.

\bibitem[Berti \& Rigo(1997)Berti and Rigo]{berti1997glivenko}
Patrizia Berti and Pietro Rigo.
\newblock A glivenko-cantelli theorem for exchangeable random variables.
\newblock \emph{Statistics \& probability letters}, 32\penalty0 (4):\penalty0 385--391, 1997.

\bibitem[Bojchevski et~al.(2020)Bojchevski, Gasteiger, and G{\"u}nnemann]{bojchevski2020efficient}
Aleksandar Bojchevski, Johannes Gasteiger, and Stephan G{\"u}nnemann.
\newblock Efficient robustness certificates for discrete data: Sparsity-aware randomized smoothing for graphs, images and more.
\newblock In \emph{International Conference on Machine Learning}, pp.\  1003--1013. PMLR, 2020.

\bibitem[Clopper \& Pearson(1934)Clopper and Pearson]{clopper1934use}
Charles~J Clopper and Egon~S Pearson.
\newblock The use of confidence or fiducial limits illustrated in the case of the binomial.
\newblock \emph{Biometrika}, 26\penalty0 (4):\penalty0 404--413, 1934.

\bibitem[Cohen et~al.(2019)Cohen, Rosenfeld, and Kolter]{cohen2019certified}
Jeremy Cohen, Elan Rosenfeld, and Zico Kolter.
\newblock Certified adversarial robustness via randomized smoothing.
\newblock In \emph{international conference on machine learning}, pp.\  1310--1320. PMLR, 2019.

\bibitem[Cresswell et~al.(2024)Cresswell, Sui, Kumar, and Vouitsis]{Cresswell2024ConformalPS}
Jesse~C. Cresswell, Yi~Sui, Bhargava Kumar, and No{\"e}l Vouitsis.
\newblock Conformal prediction sets improve human decision making.
\newblock \emph{ArXiv}, abs/2401.13744, 2024.
\newblock URL \url{https://api.semanticscholar.org/CorpusID:267211902}.

\bibitem[Deng et~al.(2009)Deng, Dong, Socher, Li, Li, and Fei-Fei]{Deng2009ImageNetAL}
Jia Deng, Wei Dong, Richard Socher, Li-Jia Li, K.~Li, and Li~Fei-Fei.
\newblock Imagenet: A large-scale hierarchical image database.
\newblock \emph{2009 IEEE Conference on Computer Vision and Pattern Recognition}, pp.\  248--255, 2009.
\newblock URL \url{https://api.semanticscholar.org/CorpusID:57246310}.

\bibitem[Gendler et~al.(2021)Gendler, Weng, Daniel, and Romano]{gendler2021adversarially}
Asaf Gendler, Tsui-Wei Weng, Luca Daniel, and Yaniv Romano.
\newblock Adversarially robust conformal prediction.
\newblock In \emph{International Conference on Learning Representations}, 2021.

\bibitem[Ghosh et~al.(2023)Ghosh, Shi, Belkhouja, Yan, Doppa, and Jones]{Ghosh2023ProbabilisticallyRC}
Subhankar Ghosh, Yuanjie Shi, Taha Belkhouja, Yan Yan, Janardhan~Rao Doppa, and Brian Jones.
\newblock Probabilistically robust conformal prediction.
\newblock In \emph{Conference on Uncertainty in Artificial Intelligence}, 2023.
\newblock URL \url{https://api.semanticscholar.org/CorpusID:260334753}.

\bibitem[Gosch et~al.(2023)Gosch, Geisler, Sturm, Charpentier, Z{\"u}gner, and G{\"u}nnemann]{gosch2023adversarial}
Lukas Gosch, Simon Geisler, Daniel Sturm, Bertrand Charpentier, Daniel Z{\"u}gner, and Stephan G{\"u}nnemann.
\newblock Adversarial training for graph neural networks: Pitfalls, solutions, and new directions.
\newblock In \emph{37th Conference on Neural Information Processing Systems (Neurips)}, 2023.

\bibitem[Guo et~al.(2017)Guo, Pleiss, Sun, and Weinberger]{Guo2017OnCO}
Chuan Guo, Geoff Pleiss, Yu~Sun, and Kilian~Q. Weinberger.
\newblock On calibration of modern neural networks.
\newblock In \emph{International Conference on Machine Learning}, 2017.
\newblock URL \url{https://api.semanticscholar.org/CorpusID:28671436}.

\bibitem[Huang et~al.(2023)Huang, Jin, Cand{\`e}s, and Leskovec]{Huang2023UncertaintyQO}
Kexin Huang, Ying Jin, Emmanuel~J. Cand{\`e}s, and Jure Leskovec.
\newblock Uncertainty quantification over graph with conformalized graph neural networks.
\newblock \emph{ArXiv}, abs/2305.14535, 2023.
\newblock URL \url{https://api.semanticscholar.org/CorpusID:258865535}.

\bibitem[Jeary et~al.(2024)Jeary, Kuipers, Hosseini, and Paoletti]{jeary2024verifiably}
Linus Jeary, Tom Kuipers, Mehran Hosseini, and Nicola Paoletti.
\newblock Verifiably robust conformal prediction, 2024.

\bibitem[Kipf \& Welling(2017)Kipf and Welling]{Kipf2017SemiSupervisedCW}
Thomas Kipf and Max Welling.
\newblock Semi-supervised classification with graph convolutional networks.
\newblock \emph{ArXiv}, abs/1609.02907, 2017.

\bibitem[Krizhevsky(2009)]{Krizhevsky2009LearningML}
Alex Krizhevsky.
\newblock Learning multiple layers of features from tiny images.
\newblock 2009.
\newblock URL \url{https://api.semanticscholar.org/CorpusID:18268744}.

\bibitem[Kumar et~al.(2020)Kumar, Levine, Feizi, and Goldstein]{kumar2020certifying}
Aounon Kumar, Alexander Levine, Soheil Feizi, and Tom Goldstein.
\newblock Certifying confidence via randomized smoothing.
\newblock \emph{Advances in Neural Information Processing Systems}, 33:\penalty0 5165--5177, 2020.

\bibitem[Lee et~al.(2019)Lee, Yuan, Chang, and Jaakkola]{lee2019tight}
Guang-He Lee, Yang Yuan, Shiyu Chang, and Tommi Jaakkola.
\newblock Tight certificates of adversarial robustness for randomly smoothed classifiers.
\newblock \emph{Advances in Neural Information Processing Systems}, 32, 2019.

\bibitem[Levine \& Feizi(2021)Levine and Feizi]{levine2021improved}
Alexander~J Levine and Soheil Feizi.
\newblock Improved, deterministic smoothing for l\_1 certified robustness.
\newblock In \emph{International Conference on Machine Learning}, pp.\  6254--6264. PMLR, 2021.

\bibitem[McCallum et~al.(2004)McCallum, Nigam, Rennie, and Seymore]{McCallum2004AutomatingTC}
Andrew McCallum, Kamal Nigam, Jason D.~M. Rennie, and Kristie Seymore.
\newblock Automating the construction of internet portals with machine learning.
\newblock \emph{Information Retrieval}, 3:\penalty0 127--163, 2004.

\bibitem[Romano et~al.(2020)Romano, Sesia, and Cand{\`e}s]{Romano2020ClassificationWV}
Yaniv Romano, Matteo Sesia, and Emmanuel~J. Cand{\`e}s.
\newblock Classification with valid and adaptive coverage.
\newblock \emph{arXiv: Methodology}, 2020.

\bibitem[Sadinle et~al.(2018)Sadinle, Lei, and Wasserman]{Sadinle2018LeastAS}
Mauricio Sadinle, Jing Lei, and Larry~A. Wasserman.
\newblock Least ambiguous set-valued classifiers with bounded error levels.
\newblock \emph{Journal of the American Statistical Association}, 114:\penalty0 223 -- 234, 2018.

\bibitem[Stutz et~al.(2021)Stutz, Cemgil, Doucet, et~al.]{stutz2021learning}
David Stutz, Ali~Taylan Cemgil, Arnaud Doucet, et~al.
\newblock Learning optimal conformal classifiers.
\newblock \emph{arXiv preprint arXiv:2110.09192}, 2021.

\bibitem[Tibshirani et~al.(2019)Tibshirani, Foygel~Barber, Candes, and Ramdas]{tibshirani2019conformal}
Ryan~J Tibshirani, Rina Foygel~Barber, Emmanuel Candes, and Aaditya Ramdas.
\newblock Conformal prediction under covariate shift.
\newblock \emph{Advances in neural information processing systems}, 32, 2019.

\bibitem[Vovk et~al.(2005)Vovk, Gammerman, and Shafer]{Vovk2005AlgorithmicLI}
Vladimir Vovk, Alexander Gammerman, and Glenn Shafer.
\newblock Algorithmic learning in a random world.
\newblock 2005.

\bibitem[Yan et~al.(2024)Yan, Romano, and Weng]{yan2024provably}
Ge~Yan, Yaniv Romano, and Tsui-Wei Weng.
\newblock Provably robust conformal prediction with improved efficiency.
\newblock \emph{arXiv preprint arXiv:2404.19651}, 2024.

\bibitem[Yang et~al.(2020)Yang, Duan, Hu, Salman, Razenshteyn, and Li]{yang2020randomized}
Greg Yang, Tony Duan, J~Edward Hu, Hadi Salman, Ilya Razenshteyn, and Jerry Li.
\newblock Randomized smoothing of all shapes and sizes.
\newblock In \emph{International Conference on Machine Learning}, pp.\  10693--10705. PMLR, 2020.

\bibitem[Zargarbashi \& Bojchevski(2024)Zargarbashi and Bojchevski]{zargarbashi2024conformal}
Soroush~H Zargarbashi and Aleksandar Bojchevski.
\newblock Conformal inductive graph neural networks.
\newblock \emph{arXiv preprint arXiv:2407.09173}, 2024.

\bibitem[Zargarbashi et~al.(2023)Zargarbashi, Antonelli, and Bojchevski]{Zargarbashi2023ConformalPS}
Soroush~H. Zargarbashi, Simone Antonelli, and Aleksandar Bojchevski.
\newblock Conformal prediction sets for graph neural networks.
\newblock In \emph{International Conference on Machine Learning}, 2023.
\newblock URL \url{https://api.semanticscholar.org/CorpusID:260927483}.

\bibitem[Zargarbashi et~al.(2024)Zargarbashi, Akhondzadeh, and Bojchevski]{zargarbashirobust}
Soroush~H Zargarbashi, Mohammad~Sadegh Akhondzadeh, and Aleksandar Bojchevski.
\newblock Robust yet efficient conformal prediction sets.
\newblock In \emph{Forty-first International Conference on Machine Learning}, 2024.

\end{thebibliography}

\newpage
\appendix

\section{Algorithm for Robust (and Vanilla) \method}
\label{sec:algorithms}

Here we provide the algorithm for \method in both $p$-fixed, and $\tau$-fixed setups. The two setups differ in calibration and finite sample correction, while computing the certificate, and returning prediction sets is similar in both. Note that in $p$-fixed version after computing the quantile $\tau_\alpha$ we correct for finite samples which results in a lower $p_\alpha^{\downarrow}$.

Note that in both algorithms we set $m$, and $\eta$ as fixed hyper-parameters defining the number of random samples per each datapoint and the collective failure probability of the confidence intervals. Therefore $\mathrm{ClopperPearson}(p)$ actually refers to the Clopper-Pearson interval with $p \cdot m$ success out of $m$ samples and failure probability of $\eta / (k + |\dcal|)$.  

\begin{algorithm}[h]
\caption{\method with $\tau$-fixed setup}
\label{alg:tau-fixed}
\SetAlgoLined
\KwIn{Score function $s: \gX \times \gY \rightarrow \sR$; Calibration set $\mathcal{D} = \{\vx_i, y_i\}_{i=1}^n$; Smoothing scheme $\xi$; Threat model $\gB$ satisfying the assumption in \autoref{thrm:robust-reduced}; Fixed threshold $\tau$, and (potentially perturbed) test point $\xtestpert$.}
\KwOut{A prediction set $\bar\gC_\mathrm{bin}(\xtestpert)$ with $1 - \alpha$ robust coverage probability}

\For{each calibration point $(\vx_i, y_i) \in \dcal$}{
    Sample from $\vx_i + \vepsilon$ for $m$ times\;
    Compute $q_i = \frac{1}{m}\sum_{j=1}^{m}\sI[s(\vx_i + \vepsilon, y_i) \geq \tau]$\;
    \eIf{Exact Certificate}{
        $q_{i}^\downarrow := q_i$\;
    }{
        $q_{i}^\downarrow := \mathrm{ClopperPearson}_\mathrm{low}(q_i)$\;
    }
}
Set $p_\alpha^\downarrow = \quantile{\alpha}{\{q_{i}^\downarrow\}_{i=0}^n}$\;
Compute $\lowerval[p_\alpha^{\downarrow}, \gB]$\ from \autoref{eq:upperbound-binary} (Lower bound minimization)\;
\For{each class $y \in \mathcal{Y}$}{
    Sample from $\xi(\xtestpert)$ for $m$ times\;
    Compute $q_{n+1,y} = \frac{1}{m}\sum_{j=1}^{m}\sI[s(\xtestpert + \vepsilon, y) \geq \tau]$\;
    \eIf{Exact Certificate}{
        $q_{n+1,y}^\uparrow := q_{n+1}$\;
    }{
        $q_{n+1,y}^\uparrow := \mathrm{ClopperPearson}_\mathrm{high}(q_{n+1, y})$\;
    }
}
\Return $\bar\gC\{ y: q_{n+1,y}^\uparrow \ge \lowerval[p_\alpha^\downarrow, \gB] \}$

\end{algorithm}

\begin{algorithm}[t]
\caption{\method with $p$-fixed setup}
\label{alg:p-fixed}
\SetAlgoLined
\KwIn{Data, score function, smoothing, and $\gB$ same as \autoref{alg:tau-fixed}. Fixed probability $p_\alpha$}
\KwOut{Same as \autoref{alg:tau-fixed}}

Update $p_\alpha \gets \lceil{p_\alpha \cdot m}\rceil/m$ (accounting for discrete samples)\;

\For{each calibration point $(\vx_i, y_i) \in \dcal$}{
    Sample from $\vx_i + \vepsilon$ for $m$ times\;
    Compute $\tau_i = \quantile{p_\alpha}{\{s(\vx_i + \vepsilon, y_i)\}_{j=1}^{m}} $\;
    \eIf{Exact Certificate}{
        $p_{\alpha}^\downarrow := p_\alpha$\;
    }{
        $p_{\alpha}^\downarrow := \mathrm{ClopperPearson}_\mathrm{low}(p_{\alpha})$\;
    }
}
Set $\tau_\alpha = \quantile{\alpha}{\{\tau_{i}\}_{i=0}^n}$\;
Compute $\lowerval[p_{\alpha}^\downarrow, \gB]$\ from \autoref{eq:upperbound-binary} (Lower bound minimization)\;
\For{each class $y \in \mathcal{Y}$}{
    Sample from $\xi(\xtestpert)$ for $m$ times\;
    Compute $q_{n+1,y}^\uparrow = \frac{1}{m}\sum_{j=1}^{m}\sI[s(\xtestpert + \vepsilon, y) \geq \tau_\alpha]$\;
    $q_{n+1,y}^{\uparrow} := $ same as \autoref{alg:tau-fixed}\;
}
\Return $\bar\gC\{ y: q_{n+1,y}^\uparrow \ge \lowerval[p_{\alpha}^\downarrow, \gB] \}$

\end{algorithm}

\parbold{Score functions} Throughout the paper we reported results for TPS score -- directly setting the softmax values as the score $s(\vx, y) = \pi(\vx, y)$ \citep{Sadinle2018LeastAS}. In vanilla CP ($r=0$), TPS tends to over-cover easy examples and under-cover hard ones \citep{Angelopoulos2021AGI}. 
Alternatively ``adaptive prediction sets'' (APS) aiming for conditional coverage uses the score function defined as $s(\vx, y):= - \left( \rho(\vx, y) + u \cdot \pi(\vx)_{y} \right)$ where $\rho(\vx,y):= \sum_{c = 1}^{K} \pi(\vx)_c 1\left[\pi(\vx)_c > \pi(\vx)_y\right]$ is the sum of all classes predicted as more likely than $y$, and $u \in [0, 1]$ is a uniform random value that breaks the ties between different scores to allow exact $1 - \alpha$ coverage \citep{Romano2020ClassificationWV}. 
Another approach is to directly use the logits of the model as the score. This is not applicable in CAS and RCSP+ \citep{zargarbashirobust, yan2024provably} as they work with bounded scores. \method can also work with unbounded score, hence, we also report the results on CP with logits. 
All three score functions are reported in \subref{fig:vanilla-l1-scores}{left} for \method. Interestingly we do not see any significant difference in set size between APS and TPS when smoothed. We report \method with APS score in \autoref{sec:additional-expr} over all $\sigma$, and radii.
Orthogonally, while RSCP with either scores quickly breaks to returning trivial sets, RSCP+ refines the score function through a biased temperature scaling. This can also be considered as another score function (or transformation over any score function) tailored for robust setup. We compare \method with RSCP+ in \autoref{fig:rscp-plus} (\autoref{sec:additional-expr}).

\section{Computing Certificate Optimization}
\label{sec:randomized-smoothing-solution}

\parbold{Canonical view} Turns out that for isotropic Gaussian and sparse smoothing, we can always attain this minimum at canonical points -- for any test point $\vx_{n+1}$ we can translate the function, and the points to the origin, and the worst case to the border of $\gB(\vx_{n+1})$. Formally, there is a pair $(\vu, \tilde\vu)$ such that $\rho_{\vu, \tilde\vu} = \rho_{\xtest, \xtestpert}$ for any $\xtest$, and $\xtestpert \in \gB(\xtest)$. Namely for the continuous ball $\gB_r$ the canonical vectors are $\vu = \boldsymbol{0}$ and $\tilde\vu=[r, 0, 0, \dots]$. For the binary $\gB_{r_a, r_d}$ we have the canonical $\vu = [0, \dots, 0, 1, \dots, 1]$ and $\tilde\vu = \boldsymbol{1} - \vu$ where $\|\vu\|_0 = r_d$ and $\|\tilde\vu\|_0 = r_a$. Intuitively it is due to the symmetry of the ball and the smoothing distribution. To avoid many notations, we again use the $\vx$, and $\tilde\vx$ in the rest of the discussion that refers to the canonical points.

To obtain an upper or lower bound (\autoref{eq:upperbound-binary} as maximization or minimization) we partition the space $\gX$ to regions where the likelihood ratio between $(\vx, \tilde\vx)$ is constant; formally $\gX = \cup_{i = 1}^k \gR_i$ where $\forall \vz \in \gR_i: \Pr[\xi(\vx) = \vz] / \Pr[\xi(\tilde\vx) = \vz] = c_i$. For any function $h$ we can find an equivalent piecewise-constant $\hat{h}$ where inside each region it is assigned to the expected value of $h$ in that region. Let $t_i =  \Pr[\xi(\vx) = \vz]$, and $\tilde t_i = \Pr[\xi(\tilde\vx) = \vz]$ then \autoref{eq:upperbound-binary} simplifies to the following linear programming \begin{equation}
    \min_{\vh \in [0, 1]^{k}}\vh^\top \tilde \vt \quad \text{s.t.} \quad \vh^\top \vt = p_\alpha
\end{equation}
Where $\vh$, $\vt$, and $\tilde\vt$ are vectors that include the values $h_i$, $t_i$, $\tilde t_i$ for each region. The optimum solution to the simplified linear programming is obtained by sorting regions based on the likelihood ratio and greedily assigning $h$ to the possible maximum in each region until the budget $\vh^\top \vt = p_\alpha$ is met. The rest of the regions are similarly assigned to zero. This problem is equivalent to fractional knapsack. For isotropic Gaussian smoothing \citet{cohen2019certified} show that the optimal solution has a closed form $\rho_\alpha = \Phi_\sigma(\Phi^{-1}_\sigma(p_\alpha) - r)$ where $\Phi_\sigma$ is the Gaussian CDF function of the Gaussian distribution with standard deviation $\sigma$. For sparse smoothing, following \citet{bojchevski2020efficient} we solve the greedy program on at most $r_a + r_d + 1$ distinct regions. The runtime is linear w.r.t. to the add and delete budget. For more detailed explanation see \citep{lee2019tight, bojchevski2020efficient}.

\section{Supplementary to Theory}
\label{sec:proofs}

\subsection{Vanilla \method}

\parbold{Conformal risk control} We use conformal risk control (CRC) \citep{angelopoulos2022conformal} to prove the coverage guarantee in \method. Here we succinctly recall it before the proof of \autoref{thrm:vote-cp}. 

\begin{theorem}[Conformal Risk Control - rephrased] 
    \label{thrm:CRC}
    Let $\lambda$ be a parameter (larger $\lambda$ yields more conservative output), and $L_i: \Lambda \to (-\infty, B]$ for $i = 1, \dots, n+1$ be exchangeable random functions. If \begin{enumerate*}[label=(\roman*)] 
    \item $L_i$s are non-increasing right-continuous w.r.t. $\lambda$,
    \item for $\lambda_\mathrm{max} = \sup \Lambda$ we have $L_i(\lambda_\mathrm{max}) \le \alpha$, and
    \item $\sup_\lambda L_i \le B < \infty$
\end{enumerate*}, then we have:
    \begin{align}
        \label{eq:lambda-hat}
        \E[L_{n+1}(\hat{\lambda})] \le \alpha \quad \text{for} \quad \hat{\lambda} = \inf \mathset{\lambda: \frac{\sum_{i=1}^n L_i(\lambda)}{n+1} + \frac{B}{n+1} \le \alpha}
    \end{align}
\end{theorem}

In case that $B = 1$, by simplifying \autoref{eq:lambda-hat}, we have $\hat{\lambda} = \inf \mathset{\lambda: \sum_{i=1}^n L_i(\lambda) \le \alpha(n+1) - 1}$
We use this framework to prove the guarantee in \method. 
\begin{proof}[Proof to \autoref{thrm:vote-cp}]
    We prove the theorem through re-parameterizing of the conservativeness variable in each case. For fixed $p$ we set $\tau = 1-\lambda$; similarly for fixed $\tau$ we set $p = -\lambda$. In both cases, the risk is defined as
    \[
    L_i(\tau, p) = 1 - \testf(\vx_i, y_i; p, \tau)
    \] which for simplicity we define $\mathrm{reject}(\vx_i, y_i; p, \tau) = 1 - \testf(\vx_i, y_i; p, \tau)$ and by definition we have $\mathrm{reject}(\vx_i, y_i; p, \tau) = \sI[\Pr[s(\vx + \vepsilon, y) < \tau] > 1 - p] = \sI[\Pr[s(\vx + \vepsilon, y) \ge \tau] < p]$. We show that the risk function satisfies the properties for a risk function feasible to the setup in \autoref{thrm:CRC}.
    \begin{enumerate}
        \item \parbold{Non-increasing to $\lambda$} In both cases the risk $L_i$ is non-inscreasing to $\lambda$; for fixed $p$ we have \begin{align*}
        \lambda_1 < \lambda_2 \Rightarrow 1 - \lambda_1 > 1 - \lambda_2 \\    
        \Rightarrow \Pr[s(\vx + \vepsilon, y) < 1 - \lambda_1] \ge \Pr[s(\vx + \vepsilon, y) < 1 - \lambda_2]  \\
        \Rightarrow \mathrm{reject}(\vx, y; p, 1 - \lambda_1) \ge \mathrm{reject}(\vx, y; p, 1 - \lambda_2)
        \end{align*}
        Now for fixed $\tau$, let $p_\vx = \Pr[s(\vx + \vepsilon, y) \ge \tau]$ then we have \begin{align*}
            \lambda_1  < \lambda_2 \Rightarrow p_1 > p_2 \text{ means that } \sI[p_\vx \le p_1] \ge \sI[p_\vx \le p_2] \\
            \Rightarrow \mathrm{reject}(\vx, y; -\lambda_1, \tau) \ge \mathrm{reject}(\vx, y; -\lambda_2, \tau) 
        \end{align*}
        Intuitively by adapting the definition of the rejection (risk) function $\mathrm{reject}(\vx_i, y_i; p, \tau) = \sI[\Pr[s(\vx + \vepsilon, y) \ge \tau] < p]$, if we increase $\lambda$ which means decreasing $p$, the chance of rejecting a label decreases. This is because, we require the same probability mass to be lower than a smaller value.
        
        \item \parbold{Right continuous} Formally the function $\mathrm{accept}$ is
        \begin{align*}
            \mathrm{accept}(\vx, y; p, \tau) = \begin{cases}
                1 & \text{if }  \Pr[s(\vx + \vepsilon, y) \ge \tau] \ge p\\
                0 & \text{otherwise}
            \end{cases}
        \end{align*}
        Across the domain (for either $p$ or $\tau$) this function has two values and it is just non-continuous in the jump between the values. For both $p$ and $\tau$ this function is left continuous due to the $\ge$ comparison. Therefore for fixed $p$ the function $\mathrm{reject}(\vx, y; p, 1 - \lambda)$ is right continuous to $\lambda = 1 - \tau$. Similar argument follows for fixed $\tau$.
        \item \parbold{Feasibility of risks less than $\alpha$} For fixed $p > 0$ if we set $\lambda = 1 - \tau$ to $\infty$ ($\tau = -\infty$), for all $\vx_i$, we have $\testf(\vx_i, y_i; p, 0) = 1$; i.e. the risk is 0 for every data. Similarly by approaching $p$ to zero in fixed $\tau$ setup, we decrease the risk to 0 for everyone. To avoid corner cases we can restrict $\tau$ to $\max s(\vx + \vepsilon, y)$ for $x\in \gX$ from above.
        \item \parbold{Limited upperbound risk} For any parameter and any input the highest possible risk is in case of rejection which is 1 ($B=1$).
    \end{enumerate}

    \parbold{Fixed $p$} The risk function $L_i(\lambda) = \mathrm{reject}(\vx_i, y_i; p, 1-\lambda)$ which means that the prediction set $\gC(\vx_i; p, 1-\lambda)$ excludes $y_{i}$. We have \[
    \E[\mathrm{reject}(\vx_{n+1}, y_{n+1}; p, 1-\hat{\lambda})] \le \alpha \ \text{for} \ \hat{\lambda} = \inf_\lambda\mathset{\lambda: \sum_{i=1}^n \mathrm{reject}(\vx_i, y_i; p, 1-\lambda) \le \alpha(n+1) - 1} 
    \]
    Setting back the ${\tau} = 1-{\lambda}$, and rewriting the expectation as a probability form, we have \[
    \Pr[y_{n+1} \in \gC(\vx_{n+1}; p, {\tau_p})] \ge 1 - \alpha \text{\ \ for \ } {\tau_p} = \sup_\tau\mathset{\tau: \sum_{i=1}^n \testf(\vx_i, y_i; p, \tau) \ge (1 - \alpha)(n+1)} 
    \]

    In the above, we used the fact that if a test fails on $\alpha(n+1) - 1$ variables among the total of $n$ variables, it passes on $n - [\alpha(n+1) - 1]$ and $(1 - \alpha)(n+1) = n - [\alpha(n+1) - 1]$.

    \parbold{Fixed $\tau$} Similarly, we define the risk function as $L_i(\lambda) = \mathrm{reject}(\vx_i, y_i; -\lambda, \tau)$. We have \[
    \E[\mathrm{reject}(\vx_{n+1}, y_{n+1}; p_\tau, \tau)] \le \alpha \quad \text{for} \quad p_\tau = \inf_\lambda\mathset{\lambda: \sum_{i=1}^n \mathrm{reject}(\vx_i, y_i; -\lambda, \tau) \le \alpha(n+1) - 1} 
    \]
\end{proof}

\begin{proof}[Proof to \autoref{thrm:views-equal}]
    The function $\testf(\vx, y; p, \tau)$ is non-increasing in both $p$ and $\tau$. Therefore the term $\sum_{i = 1}^n\testf(\vx_i, y_i; p, \tau)$ is also non-increasing in $p$ and $\tau$ and its range is the integer numbers between $0$ and $n$ (or $[n]$). For a fixed $p$, let $\tau_\alpha(p)$ be the solution to \autoref{eq:fixed-p}, then by definition it satisfies that \[
    \sum_{i = 1}^n \testf(\vx_i, y_i; p, \tau_\alpha(p)) \ge (1 - \alpha)(n+1)
    \] This implies that $p$ satisfies the same condition for $p_\alpha(\tau_\alpha(p))$. Therefore $p_\alpha(\tau_\alpha(p)) \ge p$ as $p$ is a feasible solution in \autoref{eq:fixed-tau}. The supremum search for $\tau_\alpha(p)$ directly implies that for any positive $\delta$ we have \[
    \sum_{i = 1}^n \testf(\vx_i, y_i; p, \tau_\alpha(p)) \ge \sum_{i = 1}^n \testf(\vx_i, y_i; p, \tau_\alpha(p) + \delta) - 1
    \] which intuitively means that increasing the $\tau(p)$ by any small margin fails at least in one more $\testf$ for  calibration points. Since $\sum_{i = 1}^n \testf(\vx_i, y_i; p, \tau_\alpha(p))$ is the sum of $n$ non-increasing functions, there is one index $i$ for which \[
    \testf(\vx_i, y_i; p, \tau_\alpha(p)) = 1 \quad \text{and} \quad \testf(\vx_i, y_i; p, \tau_\alpha(p) + \delta) = 0
    \] For any small positive $\delta$. Using the definition of the accept function we have \[
    \Pr[s(\vx_i + \vepsilon, y_i) \ge \tau_\alpha(p)] \ge p \quad \text{and} \quad \Pr[s(\vx_i + \vepsilon, y_i) \ge \tau_\alpha(p) + \delta] < p
    \] Due to the continuous strictly increasing CDF for $S_i$ we have $\Pr[s(\vx_i + \vepsilon, y_i) \ge \tau(p)] = p$. Therefore for any small positive $\delta$ \[
    \testf(\vx_i, y_i; p, \tau_\alpha(p)) = 1 \quad \text{and} \quad \testf(\vx_i, y_i; p + \delta, \tau_\alpha(p)) = 0
    \] which means that the accept function for $\vx_i$ fails by adding a small number to $p$. Since all other accept functions are also non-increasing we have $\sum_{i = 1}^n \testf(\vx_i, y_i; p, \tau_\alpha(p)) \le (1 - \alpha)(n+1) - 1$. This implies that $p$ is also the supremum for $\autoref{eq:fixed-p}$ with parameter $p_\alpha(\tau)$.
\end{proof}

\subsection{Robust \method}

\begin{proof}[Proof to \autoref{thrm:robust-reduced}]
    With $f_\mathrm{true}(\vx_i) = \sI[s(\vx_i, y_i) \ge \tau_\alpha]$ for true $y_i$, the calibration-time robust prediction set is defined as $\bar{\gC}_{\mathrm{cal}}(\xtestpert) = \{ p(\xtestpert, y; \tau_\alpha) \ge \quantile{\alpha}{\{\lowerval[\bar{f}_\mathrm{true}(\vx_i), \gB]\}_{i=1}^n}\}$. By definition we have $p_\alpha = \quantile{\alpha}{\{\bar{f}_\mathrm{true}(\vx)\}_{i=1}^n}\}$. Both lower bound and upper bound functions are non-decreasing. As a result, the ranks, and hence the quantile index in $\{\bar{f}_\mathrm{true}(\vx_i)\}_{i=1}^n$ and $\{\lowerval[\bar{f}_\mathrm{true}(\vx_i), \gB]\}_{i=1}^n$ are the same. Therefore, $\quantile{\alpha}{\{\lowerval[\bar{f}_\mathrm{true}(\vx_i), \gB]\}_{i=1}^n}\} = \lowerval[p_\alpha, \gB]$.

    The test-time robust prediction set is defined as $\bar{\gC}_\mathrm{test} = \{y: \upperval[\bar{f}_y(\xtestpert), \gB^{-1}] \ge p_\alpha\}$, let $\tilde{p}_y = \bar{f}_y(\xtestpert)$ then it follows \begin{align*}
        \upperval[\bar{f}_y(\xtestpert), \gB^{-1}] \ge p_\alpha \Leftrightarrow  
        \lowerval[\upperval[\bar{f}_y(\xtestpert), \gB^{-1}], \gB] \ge \lowerval[p_\alpha, \gB]\\ \Leftrightarrow
        \bar{f}_y(\xtestpert) \ge \lowerval[p_\alpha, \gB]
    \end{align*}
    By definition $\testf(\xtestpert, y; \lowerval[p_\alpha, \gB], \tau_\alpha) = \sI[ \bar{f}_y(\xtestpert) \ge \lowerval[p_\alpha, \gB]]$.
\end{proof}

In the above, we proved that \method results in a valid conformal prediction. Here we prove the validity of robust \method to adversarial data within the bounded threat model.

\begin{proof}[Proof to \autoref{thrm:gaussian-sparse-symmetric}]
For each of the mentioned smoothing schemes we have:

    \begin{figure}[t]
        \centering
        \includesvg{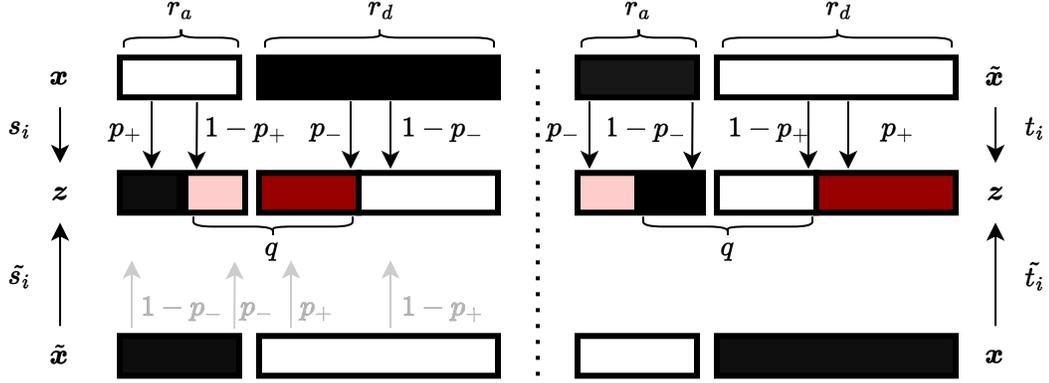}
        \caption{Illustration of likelihood ratio in sparse smoothing for both $\gB_{r_a, r_d}$, and $\gB_{r_d, r_a}$}
        \label{fig:sparse-regions}
    \end{figure}

    \parbold{Gaussian smoothing} In both cases since $p$ norm is symmetric for any point $\tilde\vx$ it holds that $\| \tilde\vx - \vx \|_p \le r$. In other words, from any perturbed point the clean point is within $\gB_r(\tilde\vx)$. Therefore $\gB^{-1}_r = \gB_r$.

    For simpler notation let $\overline{p} =\upperval[p, \gB_r]$. Given the closed from solution $\lowerval[p, \gB_r] = \Phi_\sigma(\Phi_\sigma^{-1}(p) - r)$, and $\upperval[p, \gB_r] = \Phi_\sigma(\Phi_\sigma^{-1}(p) + r)$ we have \begin{align*}
        \overline{p} = \Phi_\sigma(\Phi_\sigma^{-1}(p) + r) \Leftrightarrow 
        \Phi^{-1}_\sigma(\overline{p}) = \Phi_\sigma^{-1}(p) + r \Leftrightarrow 
        \Phi^{-1}_\sigma(\overline{p}) - r = \Phi_\sigma^{-1}(p) \\ \Leftrightarrow 
        \Phi_\sigma(\Phi^{-1}_\sigma(\overline{p}) - r) = p \Leftrightarrow 
        \lowerval[\overline{p}, \gB] = p
    \end{align*}

    \parbold{Uniform smoothing} For the uniform smoothing from \citet{levine2021improved} we have that the smooth classifier is $1/(2\lambda)$-Lipschitz continuous. From that the certified lower and upper bounds are defined as $\lowerval[p, \gB_r] = p - r\cdot1/(2\lambda)$, and $\upperval[p, \gB_r] = p + r\cdot1/(2\lambda)$ Therefore
    \[
    \lowerval[\upperval[p, \gB^{-1}_r], \gB_r] = \lowerval[p+\frac{r}{2\lambda}, \gB_r] =  p+\frac{r}{2\lambda} - \frac{r}{2\lambda} = p
    \]
    A similar argument can be applied to any certificate that directly adds (or subtracts) the Lipschitz constant of the smooth classifier to the base probability.

    \parbold{Sparse smoothing} Any $\tilde\vx \in \gB_{r_a, r_d}(\vx)$ has at most $r_a$ zero bits, and $r_d$ one bits toggled from $\vx$. By toggling those bit back we can reconstruct $\vx$. The maximum needed toggles is therefore $r_d$ zero bits and $r_a$ one bits which is the definition of $\gB_{r_d, r_a}$.

    As discussed in \autoref{sec:randomized-smoothing-solution}, canonical points for $\gB_{r_a, r_d}$ are $\vx = [0, \dots, 0, 1, \dots, 1]$ and $\tilde\vx = \boldsymbol{1} - \vx$ where $\|\vx\|_0 = r_d$ and $\|\tilde\vx\|_0 = r_a$. For $\gB^{-1}_{r_a, r_d}$ the canonical points are $\vu$, $\tilde\vu$ where $\|\vu\|_0 = r_a$. By applying a permutation over $\vu$, $\tilde\vx$ and every other point in all regions we can set $\vu = \tilde\vx$, and $\tilde\vu = \vx$. For computing both $\gB_{r_a, r_d}$, and $\gB^{-1}_{r_a, r_d}$ there are $r_a + r_d + 1$ regions of constant likelihood ratio, each including all points that have the same number of total flips from the source $\vx$, or $\vu$; formally $\gR_q = \{\vz: \| \vx - \vz \|_0 = q\}$. The same region can also defined to preserve $r_d + r_a - q$ bits from $\tilde\vx$. With $\frac{s_q}{\tilde s_q}$ as the likelihood ratio of a point $\vz$ in $\gB_{r_a, r_d}$ and $q = q_a + q_d$ as the number of changes in 1 and 0 bit, we have $s_q = (p_+)^{q_a}(1-p_+)^{r_a - q_a}(p_-)^{q_d}(1-p_-)^{r_d - q_d}$, and similarly $\tilde s_q =(p_-)^{q_a}(1-p_-)^{r_a - q_a}(p_-)^{q_d}(1-p_-)^{r_d - q_d}$. Then the likelihood ratio is simplified to 
    \begin{align}
    \frac{s_q}{\tilde s_q} = \left[\frac{p_+}{1-p_-}\right]^{q-r_d} \left[\frac{p_-}{1-p_+}\right]^{q-r_a}
    \end{align}
    As illustrated in \autoref{fig:sparse-regions} regions for $\gB_{r_a, r_d}^{-1}$ are same as $\gB_{r_a, r_d}$ only with reverse order. In other word, let $t_i$, $\tilde t_i$ be the probability of visiting region $\gR_q$ from $\vu$ and $\tilde\vu$, then $t_i = \tilde s_{r_a + r_d + 1 - q}$, and $\tilde t_i = s_{r_a + r_d + 1 - q}$. For a fixed $\vz$ the probability to visit $\vz$ from $\vx$ is the probability of toggling $q = \|\vz - \vx\|_0$ bits which is the same as toggling $q$ bits from $\tilde\vu$ as $\tilde\vu = \vx$. 

    Solutions to $\lowerval[p, \gB_{r_a, r_d}]$ and $\upperval[p, \gB^{-1}_{r_a, r_d}]$ are obtained from the following optimization functions:
    \begin{align*}
    \lowerval[p, \gB_{r_a, r_d}] = \min_{\vh \in [0, 1]^{r_a + r_d + 1}}\vh^\top \tilde \vt \quad \text{s.t.} \quad \vh^\top \vt = p \\
    \upperval[p, \gB^{-1}_{r_a, r_d}] = \max_{\vh \in [0, 1]^{r_a + r_d + 1}}\vh^\top \tilde \vs \quad \text{s.t.} \quad \vh^\top \vs = p 
    \end{align*}
    The solution to the lower bound optimization is obtained by a greedy algorithm. We visit each in increasing order w.r.t. $\frac{s_q}{\tilde s_q}$, we assign $h_q = 1$ until the budget $\vh^\top \vs$ is met and we set $h_q=0$ for the remaining regions (fractional knapsack problem). For the maximization we do the same but in a decreasing order.

    We want to prove $\lowerval[\upperval[p, \gB_{r_a, r_d}^{-1}], \gB_{r_a, r_d}] = p$. This is the solution to \begin{align*}
        \min_{\vh \in [0, 1]^{r_a + r_d + 1}}\vh^\top \tilde \vt \quad \text{s.t.} \quad \vh^\top \vt = \upperval[p, \gB_{r_a, r_d}^{-1}] = \vh'^\top\tilde\vs
    \end{align*}
    Let $\overleftarrow{\tilde\vs}$ be the vector $\tilde\vs$ in reverse order. Then $\vt = \overleftarrow{\tilde\vs}$. From the problem definition we have that $\overleftarrow{\vh'}$ (the solution from maximization problem in reverse order) is a feasible solution. Given that the solution of the optimization (reduced to fractional knapsack problem) is always in form of $[1, \dots, 1, \delta, 0, \dots, 0]$ for some $\delta \in [0, 1]$, any vector of this form that satisfies the constraint is optimal. Therefore $\overleftarrow{\vh'}$ is the solution to the maximization greedy problem. So the optimal solution is $\overleftarrow{\vh'}^\top\tilde\vt = \overleftarrow{\vh'}^\top\overleftarrow{\vs} = p$.
\end{proof}

For any $\ell_p$ with the same argument as $\ell_2$ ball we have $\gB^{-1}_r = \gB_r$. Similar to isotropic Gaussian smoothing, the Lipschitz continuity in DSSN-smoothed distribution shows that \autoref{thrm:gaussian-sparse-symmetric} applies to $\ell_1$ ball and this distribution as well.

\subsection{Correction for Finite Sample Monte-Carlo Estimation}
\label{sec:finite-sample-more}

\begin{proof}[Proof to \autoref{thrm:MC-tau-fixed}]
    With $p_i = \Pr[s(\vx_i + \vepsilon, y_i) \ge \tau_\alpha]$ as the true probability of crossing $\tau_\alpha$ for each true score distribution in calibration set. We have $p_\alpha = \quantile{\alpha}{\{p_i\}_{i=1}^{n}}$. For all $i$ we have $q_{i}^\downarrow \le p_i$ from which follows $p_{\alpha}^\downarrow \le p_\alpha$.
    The probability of failure in each calibration datapoint is $\eta/(|\dcal| + k)$; as a result, from the union bound the probability of failure $q_{i}^\downarrow \le p_i$ for all $i \in \{1, ..., n\}$ and therefore the quantile is $|\dcal| \eta/(|\dcal| + k)$.
    
    For all classes of the test point we have $\tilde q_{n+1, y}^\uparrow \ge \tilde p_{n+1, y}$ with $\eta/(|\dcal| + k)$. Therefore, for the true class we have $\tilde q_{n+1} \ge \tilde p_{n+1}$ with $k\eta/(|\dcal| + k)$. 
    
    Conformal guarantee implies that with $1 - \alpha + \eta$ probability we have $p_{n+1} \ge p_\alpha$. The robustness certificate implies that $\tilde p_{n+1} \ge \lowerval[p_\alpha, \gB]$. 
    Following holds by using the mentioned inequality: \begin{align*}
         \tilde{q}_{n+1}^\uparrow\underset{1 - \frac{k\eta}{n + k}}{\ge} \tilde p_{n+1} \underset{1 - \alpha + \eta}{\ge} \lowerval[p_\alpha, \gB] \underset{1 - \frac{n\eta}{n+k}}{\ge} \lowerval[p_{\alpha}^\downarrow, \gB]
    \end{align*}
    From the union bound it follows that the total failure probability is less than $\alpha$.    
\end{proof}

We estimate the probabilities in \autoref{alg:tau-fixed} following \autoref{thrm:MC-tau-fixed} directly. There, we estimate the $p_i = \Pr_{\vepsilon}[s(\vx_i + \vepsilon, y_i) \ge \tau]$ via Monte Carlo samples resulting in $q_i = \frac{1}{m}\sI[s(\vx_i + \vepsilon, y_i) \ge \tau]$. Via Clopper Pearson bounds we find $q_i^\downarrow$ for which we have $q_i^\downarrow \le p_i$ with adjusted $1 - \eta$ probability.

For the $p$-fixed approach (\autoref{alg:p-fixed}),  we compute the discrete quantile $\tau_i = \sQ(\lceil p_\alpha\cdot m\rceil / m;\{s(\vx_i + \vepsilon, y_i)\}_{i = 1}^m$ over the randomly sampled scores. By definition of discrete quantile function, without counting again, we have $\frac{1}{m}\sI[s(\vx_i + \vepsilon, y_i) \ge \tau_i] \ge p_\alpha$ ($p_\alpha$ proportion of the binary variables $\sI[s(\vx_i + \vepsilon, y_i) \ge \tau_i]$ are 1). Now, we bound a Bernoulli parameter that is the proportion of the distribution exceeding $\tau_i$. Again we use Clopper Pearson bound but this time on $p_\alpha$ which is known. Therefore after computing $\tau_\alpha$ we use $p_\alpha^\downarrow$ instead which is a minimum support for all $\tau_i$-s set as the $p_\alpha$ quantile of the sampled scores (with adjusted $1 - \eta$ confidence).

\begin{figure}
    \centering
    \input{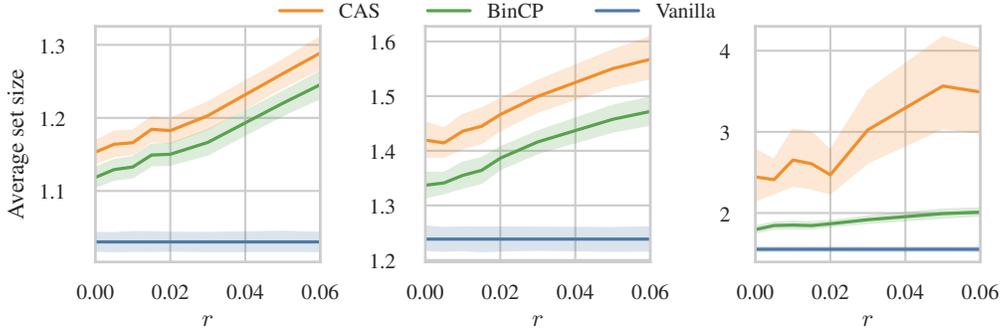}
    \caption{Comparison between \method and CAS on CIFAR-10 dataset with $\sigma=0.25$ and small values of $r$. The nominal coverage $1 - \alpha$ is set to [From left to right] 85\%, 90\%, and 95\%.}
    \label{fig:low-r}
\end{figure}

\section{Supplementary Discussion}
\label{sec:supplementary-discussion}

\subsection{High-level understanding of robustness certificates} 
\label{sec:suppl-understanding-certificates}

A certificate of robustness is a formal guarantee that the model predicts the same class for any perturbation within the specified threat model - within the ball around the input. In other words, if the function $f$ is certified to be robust for the point $\vx$ w.r.t. $\gB$, for any $\tilde\vx \in \gB(\vx)$ we have $\arg\max_y f_y(\vx) = \arg\max_y f_y(\tilde\vx)$.  This certificate ensures that the top label remains the same within the threat model (binary certificate). A similar (confidence, or soft) certificate guarantees a lower (and upper) bound on the model confidence within the threat model given the predictive probability.
One way to attain such certificates is through verifiers. Verifiers need white-box access (knowledge about the model structure and weights) and they work efficiently only on a limited class of models. However, our robust conformal guarantee is black-box. 

A common approach for black-box certification is through randomized smoothing. A randomly smoothed classifier results from inference given the input augmented with random noise. For example $g(\vx) = \E_{\vepsilon \sim \gN(\boldsymbol{0}, \sigma^2\boldsymbol{I})}[f(\vx + \vepsilon)]$ -- model $g$ returns the expected output of $f$ given randomly augmented $\vx$ where the noise comes from an isotropic Gaussian distribution with scale $\sigma$. The randomization function is smooth even if the original function changes rapidly, which is the effect of the expectation. It is also Lipschitz continuous, meaning that we can bound the output based on the distance of $\tilde\vx$ from $\vx$.
The latter allows us to provide formal guarantees that the top class probability (or confidence) remains high (changes slowly) even if $\tilde\vx \in \gB$ is passed to the model instead of $\vx$.

Ultimately a randomized smoothing-based certificate returns a lower (or upper) bound probability (or score) on the expected output (given the randomized $\vx$). In robust CP we use these bounds to answer ``if instead of the clean input $\xtest$ which is already exchangeable with the calibration set, the model received the worst case $\xtestpert \in \gB(\xtest)$ how much lower the conformity score has become''. Or in other words ``if the model is queried with $\xtestpert \in \gB(\xtest)$ (which has a lower conformity score in order not to be covered) how much higher the conformity score of the clean input can be''. Technical details of smoothing-based certificates are mentioned in \autoref{sec:robust-votecp}. For a more detailed discussion see \citep{cohen2019certified, kumar2020certifying}.

\begin{figure}
    \centering
    \input{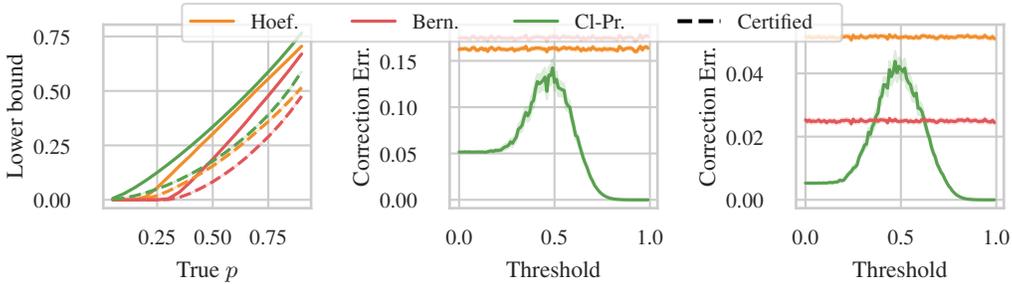}
    \caption{[Left] Confidence lower bound and the corresponding certified lower bound for scores derived from Beta and Bernoulli distributions.  [Middle and right] Correction error (lower bound subtracted from the theoretical mean) of the scores distributed from the Gaussian distribution both in continuous case (mean lower bound) and binarized case (lower bound on the Bernoulli parameter) for [Middle] 100 and [Right] 1000 samples. Details of the experiments are in \autoref{sec:suppl-confidence-intervals}.}
    \label{fig:conf-radius}
\end{figure}

\subsection{Comparison of confidence intervals}
\label{sec:suppl-confidence-intervals}
As discussed in \autoref{sec:monte-carlo}, \method, CAS, and RSCP, all require true probability, CDF, and mean from the distribution of scores which is intractable to compute (except in the case of de-randomized \method). Therefore we use confidence bounds that are lower (or higher) than the true values with collective probability $1 - \eta$ (which is taken into account while calibrating). CAS, and RSCP are defined on continuous scores that are bounded by Hoeffding, Bernstein, or DKW inequalities. \method is defined through binarized scores, and the final parameter is the success probability of a Bernoulli distribution which can bounded by the Clopper-Pearson interval which is exact \citep{clopper1934use}. The width of all mentioned confidence intervals is decreasing w.r.t. the sample size. Therefore a tighter interval can result in the same or better efficiency (correction error) with fewer samples; e.g. For scores sampled from a Gaussian $\gN(0.5, 0.1)$ Clopper Pearson error (for $z\ge0.6$) with 100 samples is still lower than Bernstein's error with 250 samples.

To illustrate this we conducted two experiments. First, to compare the tightness of each concentration inequality, we sampled from a Beta distribution with mean $p$ to have continuous score values between $[0, 1]$. The distribution for a fixed $\beta$ is $\mathrm{Beta}(\frac{p}{\beta(1 - p)}, \beta)$. Then for the continuous score, we computed both Hoeffding's and Bernstein's lower bound on the mean, alongside the Clopper-Pearson bound for the given parameter $p$ and the same sample size. As shown in \subref{fig:conf-radius}{left} (with $\beta=1$) the binary lower bound is always higher (better). Since the certified lower bound is an increasing function of the given probability, the certified lower bound for the binary values is again higher. \looseness=-1

In another experiment shown in \subref{fig:conf-radius}{middle and right} we sample scores from a Gaussian distribution $\gN(0.5, 0.1)$, and computed the lower-bound mean given both Hoeffding and Bernstein's inequalities. Then for various thresholds, we computed the probability of scores passing that threshold and lower bounded this probability by Clopper Pearson concentration inequality. As shown in the figure for lower sample rates, binarization results in less error compared to the theoretical mean. Even with higher sample rates Clopper Pearson interval is significantly tighter than the other two for low and high thresholds (which is a parameter of \method).

\begin{proposition}
\label{thrm:clopper-better-hoeff}
    Let $X \sim \mathrm{Beta}(a, b)$ and $x_1, \dots, x_m$ be $m$ i.i.d. samples of $X$. Given the empirical mean $\bar{x} = \frac{1}{m}\sum_{i = 1}^{m}x_i$ the upper bound for the true mean $\mu = \E[X]$ is given by the Hoeffding's inequality as $\mu \le \bar{x} + b_\mathrm{hoef}$, where $b_\mathrm{hoef} = \sqrt{\frac{\ln(\frac{1}{\eta})}{2m}}$.  For any user-specified $\tau \in (0, 1)$, let $Y = \sI[X > \tau]$. The Clopper-Pearson upper bound $p_u$ for the true $p = \E[Y] = \Pr[X \ge \tau]$ is:
    \begin{align*}
        p_u = \Phi^{-1}_\mathrm{Beta}(1 - \eta; 1+\sum_{i = 1}^{m}\sI[x_m > \tau], m - \sum_{i = 1}^{m}\sI[x_m > \tau_\mu])
    \end{align*}
    Each upper bound holds with probability $1 - \eta$.
    For any number of samples $m$, and any significance level $\eta$, the probability that the CP bound is tighther is  $\Pr[p_u - \E[Y] \le b_\mathrm{hoef}]$ and equals:
    \begin{align}
        \Pr[p_u - \E[Y] \le b_\mathrm{hoef}] = \Phi_\mathrm{Binom}(\hat{m}; m, \E[Y])
    \end{align} where $\hat{m}$ is defined in \autoref{eq:m-overlap-hoef}.
\end{proposition}

\begin{figure}[t]
    \centering
    \input{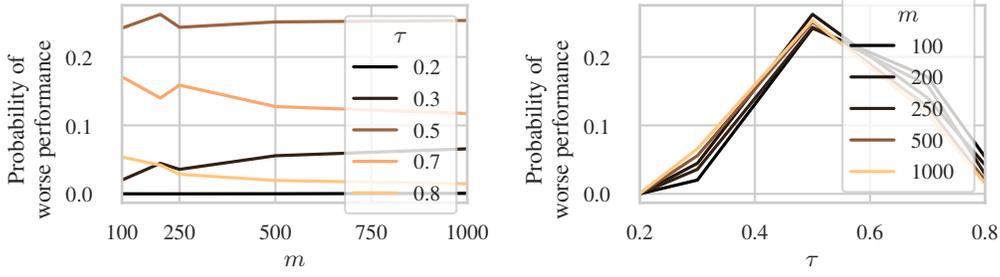}
    \caption{Probability of observing higher upper bound from Clopper Pearson confidence interval in comparison with Hoeffding's interval. The result is for $\mathrm{Beta}(2, 2)$, and $\eta = 0.01$.}
    \label{fig:hoef-clop}
\end{figure}

\begin{proof}
    The variable $Y$ is distributed as $Y \sim \mathrm{Bernoulli}(p)$ where $p = \E[Y] = 1 - \Phi_\mathrm{Beta}(\tau; a, b)$ and $\Phi_\mathrm{Beta}$ is the CDF of the beta distribution with parameters $a$ and $b$.

    Let $m_+ = \sum_{i = 1}^{m}\sI[x_m > \tau]$. We will compute the probability that the inequality
    \begin{align}
        p_u - \E[Y] \le \sqrt{\frac{\ln(\frac{1}{\eta})}{2m}}
    \end{align}
    holds. Substituting the definition of $p_u$ and $\E[Y]$ we get:
    \begin{align*}
        \Phi^{-1}_\mathrm{Beta}(1 - \eta; 1+m_+, m - m_+) - (1 - \Phi_\mathrm{Beta}(\tau; a, b)) \le \sqrt{\frac{\ln(\frac{1}{\eta})}{2m}} \Leftrightarrow\\
        1 - \eta \le \Phi_{\mathrm{Beta}}\left( \sqrt{\frac{\ln(\frac{1}{\eta})}{2m}} + 1 - \Phi_\mathrm{Beta}(\tau; a, b); 1+m_+, m-m_+ \right) \Leftrightarrow \\
        \eta \ge 1 - \Phi_{\mathrm{Beta}}\left( \sqrt{\frac{\ln(\frac{1}{\eta})}{2m}} + 1 - \Phi_\mathrm{Beta}(\tau; a, b); 1+m_+, m-m_+ \right)
    \end{align*}
    Define $\hat{m}$ as the break-point after which the Clopper-Pearson bound becomes looser than the Hoeffding bound:
    \begin{align} 
    \label{eq:m-overlap-hoef}
    \hat{m} = \sup\left\{ m_+: \sI\left[\eta \ge 1 - \Phi_{\mathrm{Beta}}\left( \sqrt{\frac{\ln(\frac{1}{\eta})}{2m}} + 1 - \Phi_\mathrm{Beta}(\tau; a, b); 1+m_+, m-m_+ \right)\right] \right\}
    \end{align}
    In other words, $m_+ > \hat{m} \Leftrightarrow p_u - \E[E] > b_\mathrm{hoef}$. Since $\Phi$ is monotonic, it follows that:
    \begin{align}
        \label{eq:prob_cp_better}
        \Pr[p_u - \E[Y] > b_\mathrm{hoef}] = \Pr[m_+ > \hat{m}] = 1 - \Phi_\mathrm{Binom}(\hat{m}; m, \E[Y])
    \end{align} Where $\Phi_\mathrm{Binom}$ is the CDF of the Binomial distribution with the specified parameters.
    
\end{proof}

Similarly, we can compare Clopper-Pearson bound with the Bernstein bound we use $\mu \le \bar{x} + b_\mathrm{bern}$ where
\begin{align*}
    b_\mathrm{bern} = \sqrt{2\sigma^{2}_m\frac{\ln(\frac{2}{\eta})}{m}} + \frac{7\ln(\frac{2}{\eta})}{3(m-1)}
\end{align*}
By replacing $b_\mathrm{hoef}$ with $b_\mathrm{bern}$ in 
\autoref{thrm:clopper-better-hoeff} we can derive a similar result.
We choose a Beta distribution to simulate the fact that conformity scores such as TPS and APS are bounded. Moreover, we need bounded scores to be able to apply Hoeffding's inequality. Any other distribution (after some transformation that ensures bounded scores) could be used, as long as we can compute its CDF. 

In \autoref{fig:hoef-clop} we show the probability of Hoeffding bound being tighter than Clopper-Pearson bound (it's complement is defined in \autoref{eq:prob_cp_better}) for $X \sim \mathrm{Beta}(2, 2)$ for different values of $m$ and $\tau$. There is a choice of $\tau$ such that the probability is effectively $0$ for all values of $m$, i.e. the CP bound is always better. Interestingly, at worst, both in terms of the number of samples $m$ and $\tau$, we see that it is less than 25\%. That is, Clopper-Pearson bound is better on average for all configurations.

To get some additional intuition, instead of the exact Clopper-Pearson bound for $p$ we can use the following bound derived from a Normal approximation which approximately holds with probability $1-\eta$:
\begin{align*}
    p \leq \hat{p} + \frac{z_\eta}{\sqrt{m}}\sqrt{\hat{p}(1 - \hat{p})}
\end{align*} where $\hat{p}=\frac{1}{m}\sum_{i = 1}^{m}\sI[x_m > \tau]$ and
$z_\eta$ is the $1 - \eta$ quantile of the standard normal distribution. 
It is not difficult to verify that for all values of $\hat{p} \in [0, 1]$
we have that 
\[
 \frac{z_\eta}{\sqrt{m}}\sqrt{\hat{p}(1 - \hat{p})} \le \sqrt{\frac{\ln(\frac{1}{\eta})}{2m}}
\]
To see this, note that the $\sqrt{m}$ term cancels, and for $\eta=0.05$ $z_\eta \approx 1.64, \sqrt{\frac{\ln(\frac{1}{\eta})}{2}} \approx 1.22$. Since  $\sqrt{\hat{p}(1 - \hat{p})} \in [0, 0.5]$, even in the worst-case $1.64 \cdot 0.5 \le 1.22$. This analysis again confirms that CP gives tighter bounds.
\autoref{thrm:clopper-better-hoeff} can be analogues extended to lower bounds.



\begin{figure}[t]
    \centering
    \input{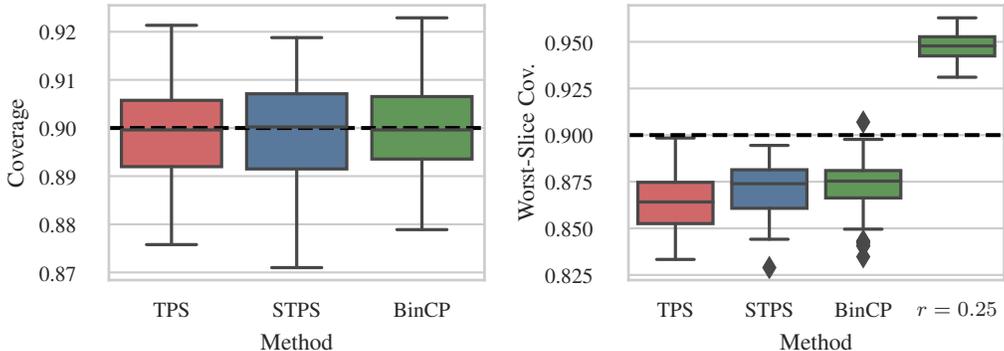}
    \caption{Comparison of coverage [Left] and worst-slice coverage [Right]. Here the STPS refers to the smooth TPS which is the average of 2000 randomly smooth inferences per point. The results are for CIFAR-10 dataset and r= 0 unless specified.}
    \label{fig:conditional}
\end{figure}

\parbold{Finite sample correction for fixed $\tau$ setup}
What we showed in \autoref{thrm:MC-tau-fixed} adds MC sample correction to \method with fixed $\tau$ computation. We can correct for finite samples in a fixed $p$ setup in a similar way. First, we compute the $\tau_i(p)$ for each of the calibration points. In an asymptomatically valid setup this implies that for $\tau_i(p)$ we have $\Pr[s(\vx_i + \vepsilon, y_i) \ge \tau_i(p)] \ge p$. To account for finite samples we reduce $p$ to $p^\downarrow$ ($p^\downarrow\le p$). Again this holds for each calibration with $1/(|\dcal| + k)$ probability, and the conformal threshold is $(\tau_p, p^\downarrow)$. In the test time the setup is identical to the fixed-$\tau$ setup. 

\subsection{Realistic setup to evaluate GNN robustness} 
\label{sec:suppl-gnn-setup}

Concurrent to our work \citet{gosch2023adversarial} shows that transductive setup is flawed to evaluate GNN robustness. This is since assuming that the clean graph is accessible to the defender during training, can lead to perfect robustness by simply remembering the clean graph. Many robust- and self-training GNNs proposed for robustness also exploit this flaw. 

To evaluate GNN node-classification for robust CP, following prior works, we assumed the transductive setup where the clean graph is given during the training and calibration. Specifically, we assume that the defender trains, and computes the calibration scores on the clean graph. Perturbations in test nodes are then applied after calibration (in evasion setup). This again is based on similar assumption made by many other GNN robustness works. Therefore our robust node-classification results are only representing the comparison of \method and other approaches on a sparse binary certificate.

A more realistic setup is the inductive setup where the defender is given a clean subgraph $\gG_\mathrm{tr}$ for training (and calibration), and the adversary perturbs the rest of the graph upon arrival; i.e. the defender does not know the clean test graph. In vanilla setup, the conformal guarantee is still valid in the inductive setting via re-computing calibration nodes \citep{zargarbashi2024conformal}. However in evasion, robust CP is not as easily applicable by recomputing the quantile and computing upper bounds on scores, since the calibration nodes are also affected by the adversarial test nodes through the message passing. Therefore robustness in GNNs under realistic setups is challenging which we leave it to future works.

\section{Additional Experiments}
\label{sec:additional-expr}

\parbold{Various models and smoothing maginitudes ($\sigma$)}
In \autoref{tab:all-sigmas} we compare the result between SOTA CAS and \method for CIFAR-10 dataset. The results are reported across various data smoothing $\sigma$ values, and models trained with different noise augmentations (data augmented during training with different $\sigma$ values). We call them smoothing $\sigma$, and model $\sigma$ respectively. In the robustness certificate for classification, it is considered best practice to use the same $\sigma$ in both model's noise augmentation, and the smoothing process. Similarly, in robust conformal prediction mismatching smoothing and model $\sigma$ results in a larger prediction set. Interestingly this adverse effect is much less observed in \method although it remains present. Overall, across all smoothing parameters, model $\sigma$ values, coverage rates, and perturbation radii, \method consistently outperforms CAS.

\parbold{Performance on small radii} For completeness, in \autoref{fig:low-r}, we report the performance of \method on small values of $r$. As \citet{jeary2024verifiably} reports $\sim 4.45$ average set size for $r=0.02$ (Table 1 in \citep{jeary2024verifiably}) our report shows more than twice smaller sets for the same $r$. As in \autoref{tab:all-sigmas} we observe the same average set size for $r\sim 0.5$ ($\ge 20\times$ higher radius) for smallest $\sigma=0.12$. As we discussed, one effect of this eye-catching difference is the inherent robustness of the randomized smooth prediction. As shown in \autoref{fig:vanilla-l1-scores}, the empirical coverage of non-smooth prediction drastically decreases to 0 for small radii, while in smooth prediction the coverage decreases slowly. 

\begin{table}[htbp]
\centering
\caption{Comparison of smoothing-based robust CP methods on APS score}
\label{tab:aps-table}
\resizebox{\textwidth}{!}{
\begin{tabular}{cccccccccc}
\toprule
\multicolumn{2}{c}{} & \multicolumn{4}{c}{$1 - \alpha=0.9$} & \multicolumn{4}{c}{$1 - \alpha = 0.95$} \\
 &  & \multicolumn{2}{c}{CAS} & \multicolumn{2}{c}{\method} & \multicolumn{2}{c}{CAS} & \multicolumn{2}{c}{\method} \\
\cmidrule{3-10}
\textbf{$\sigma$} & \textbf{r} & \textbf{Coverage} & \textbf{Set Size} & \textbf{Coverage} & \textbf{Set Size} & \textbf{Coverage} & \textbf{Set Size} & \textbf{Coverage} & \textbf{Set Size} \\
\midrule

\multirow{7}{*}{0.12} & 0.06 & 0.954 & 1.635 & 0.946 & 1.529 & 0.990 & 4.022 & 0.980 & 2.151 \\
 & 0.12 & 0.971 & 1.939 & 0.963 & 1.757 & 0.996 & 6.435 & 0.985 & 2.389 \\
 & 0.18 & 0.987 & 2.879 & 0.978 & 2.076 & 1.000 & 9.745 & 0.991 & 2.876 \\
 & 0.25 & 0.998 & 7.454 & 0.986 & 2.510 & 1.000 & 10.000 & 0.995 & 3.405 \\
 & 0.37 & 1.000 & 10.000 & 1.000 & 10.000 & 1.000 & 10.000 & 1.000 & 10.000 \\
 & 0.50 & 1.000 & 10.000 & 1.000 & 10.000 & 1.000 & 10.000 & 1.000 & 10.000 \\
 & 0.75 & 1.000 & 10.000 & 1.000 & 10.000 & 1.000 & 10.000 & 1.000 & 10.000 \\

\midrule

\multirow{7}{*}{0.25} & 0.06 & 0.955 & 2.108 & 0.944 & 1.894 & 0.986 & 3.316 & 0.976 & 2.677 \\
 & 0.12 & 0.964 & 2.309 & 0.954 & 2.054 & 0.989 & 3.682 & 0.980 & 2.857 \\
 & 0.18 & 0.970 & 2.495 & 0.961 & 2.227 & 0.993 & 5.038 & 0.986 & 3.181 \\
 & 0.25 & 0.980 & 2.900 & 0.972 & 2.537 & 0.997 & 6.004 & 0.989 & 3.444 \\
 & 0.37 & 0.991 & 3.795 & 0.982 & 3.047 & 1.000 & 9.360 & 0.994 & 4.035 \\
 & 0.50 & 0.999 & 7.430 & 0.991 & 3.729 & 1.000 & 10.000 & 0.997 & 4.850 \\
 & 0.75 & 1.000 & 10.000 & 1.000 & 10.000 & 1.000 & 10.000 & 1.000 & 10.000 \\

\midrule

\multirow{7}{*}{0.50} & 0.06 & 0.956 & 2.738 & 0.942 & 2.479 & 0.981 & 3.864 & 0.975 & 3.342 \\
 & 0.12 & 0.962 & 2.890 & 0.951 & 2.635 & 0.984 & 4.077 & 0.978 & 3.508 \\
 & 0.18 & 0.968 & 3.078 & 0.959 & 2.801 & 0.986 & 4.277 & 0.981 & 3.658 \\
 & 0.25 & 0.973 & 3.304 & 0.966 & 2.994 & 0.989 & 4.546 & 0.984 & 3.899 \\
 & 0.37 & 0.980 & 3.684 & 0.974 & 3.302 & 0.993 & 5.193 & 0.988 & 4.300 \\
 & 0.50 & 0.986 & 4.153 & 0.979 & 3.663 & 0.996 & 5.868 & 0.991 & 4.733 \\
 & 0.75 & 0.995 & 5.441 & 0.989 & 4.584 & 0.999 & 8.026 & 0.995 & 5.542 \\

\bottomrule
\end{tabular}}
\end{table}

\parbold{APS score function} Although we observe similar comparison between CAS and \method given APS score function, for completeness we report the performance of both methods in \autoref{tab:aps-table}.

\parbold{Conditional and class-conditional coverage} Following \citet{Romano2020ClassificationWV}, we approximated the conditional coverage gap as the worst coverage among $n$ different slices. Each slice is defined as $\gX_s=\mathset{\vx_i \in \gD: a\le \vx_i \cdot \vv\le b}$ for the random vector $\vv$ and random scalers $a, b$ \citep{Romano2020ClassificationWV}. For that, we sampled 200 random vectors $\vv$ and among all the scalers randomly sampled $a, b$ from the set  $\mathset{\vx_i\cdot\vv, \vx_i \in \gD}$. We report the result over 100 different calibration samplings. In each iteration of the experiment, we exclude the slices with less than 200 points of support. 

\method has smoothing, binarization and accounting for perturbation radius. For a better intuition we observe the effect of each step separately: vanilla TPS (without smoothing) as the baseline score, vanilla smooth TPS (labeled as STPS) which is an average of TPS scores over randomized sampled (same as CAS), \method without robustness (set $r=0$) which reflects the effect of binarization (we did not correct for finite sample due to validity in vanilla setup), and robust CP via \method. 
As shown in \autoref{fig:conditional}, the smooth model has a better worst-slice coverage than vanilla TPS. Though binarization although the average worst-slice coverage remains the same, there is a slight decrease in the variance of this metric. 
Note that sampling correction and making CP robust increases the empirical coverage guarantee, therefore the worst slice coverage is increased due to the inherent increase in marginal coverage.

\begin{figure}[ht]
    \centering
    \input{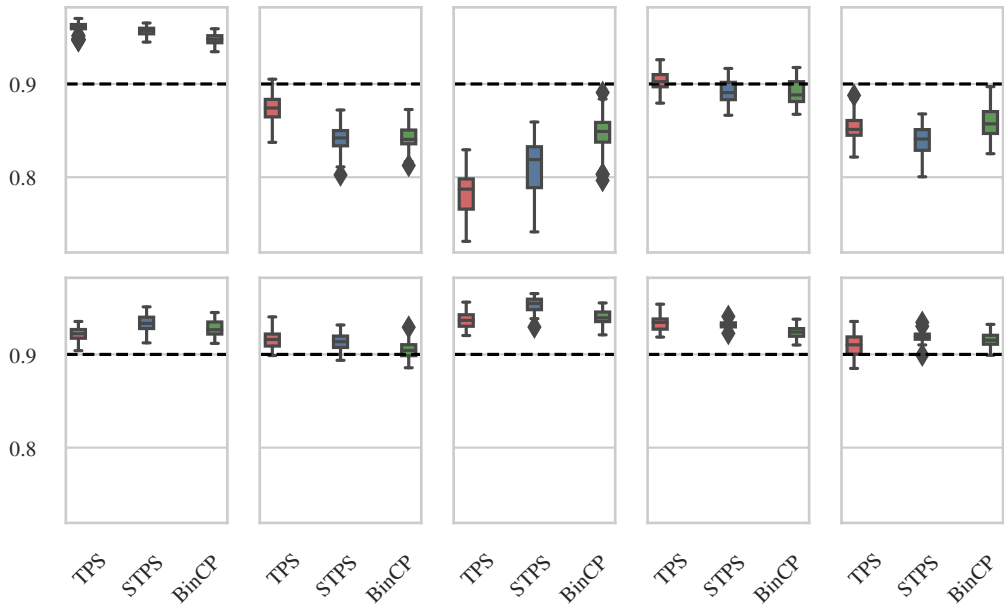}
    \caption{Comparison of methods in class-conditional coverage for all classes of CIFAR-10, note that here \method is used without sampling correction. That is because the correction slightly increases empirical coverage which can be misleading.}
    \label{fig:class-conditional}
\end{figure}
 
 We also reported the result of the class-conditional coverage in \autoref{fig:class-conditional}. Empirically in almost all classes, \method is closer to the nominal guarantee compared to normal smoothing. Ultimately both smooth prediction and \method are not comparable with vanilla TPS.

\begin{figure}
    \centering
    \input{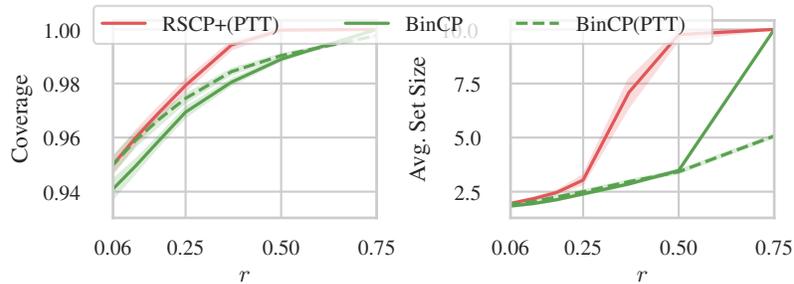}
    \caption{Comparison between \method and RSCP+ (PPT, \autoref{eq:ppt-score}) and \method with \autoref{eq:ppt-score}. The result is on CIFAR-10 dataset with $\sigma=0.25$.}
    \label{fig:rscp-plus}
\end{figure}

\parbold{Comparison with RSCP+}
\citet{yan2024provably} shows a flaw of RSCP \citep{gendler2021adversarially} indicating that the score function is not corrected for finite sample estimation. They show that by adding finite sample correction to RSCP, it becomes significantly inefficient and produces trivial sets $\gC(\xtest) = \gY$. They remedy that by designing a ranking-based transformation on top of the given score function which defines a new score as 
\begin{align}
\label{eq:ppt-score}
    s_\mathrm{ppt}(\vx, y) = \sigma \left(\frac{1}{T|\gD_{\mathrm{tune}}|} \mathrm{rank}(s(\vx, y); \mathset{s(\vx_j, y_j)}_{(\vx_j, y_j) \in \gD_{\mathrm{tune}}}) - \frac{b}{T} \right)
\end{align} Where $\gD_{\mathrm{tune}}$ is a holdout tuning index, $T$ is the temperature parameter, $b$ is a bias parameter, and $\sigma$ is the sigmoid function. The original experiment from \citet{yan2024provably} has several issues, which we resolved and compared with it: \begin{enumerate*}[label=(\roman*)]
    \item The scores have possible ties; i.e. two different data points can have the same score value. To remedy that we added an unnoticeable random number $\delta \sim \mathrm{Uniform}[0, 1/|\gD_{\mathrm{tune}}|]$ to the scores.
    \item The tuning set and the calibration set in the experiments are significantly large. \citet{yan2024provably} use 5250/10000 test datapoints as a tuning and calibration set. This unrealistic holdout labeled set contradicts the sparse labeling assumption. In our reproduction of their results we used a total of $\sim 380$ datapoints where 200 of them are for tuning. 
\end{enumerate*} As shown in \autoref{fig:rscp-plus}, \method still outperforms RSCP+(PPT). As the score function in \autoref{eq:ppt-score} is also a valid score, we can use \method on top which shows slightly better efficiency for larger radii compared to \method combined with TPS score. Here we set $b=1 - \alpha$, and $T=0.001$, and report the results on 2000 MC samples. The reported result is on CIFAR-10 dataset.

\begin{table}[]
\resizebox{\textwidth}{!}{
\begin{tabular}{@{}ccccccccccc@{}}
\toprule
      &           &      & \multicolumn{4}{c}{$1 - \alpha=0.9$}                            & \multicolumn{4}{c}{$1 - \alpha=0.95$}                           \\ 
      &           &      & \multicolumn{2}{c}{\textbf{CAS}} & \multicolumn{2}{c}{\textbf{\method}} & \multicolumn{2}{c}{\textbf{CAS}} & \multicolumn{2}{c}{\textbf{\method}} \\
\textbf{Model} $\sigma$ & \textbf{Data} $\sigma$ & $r$    & Coverage & Set Size & Coverage  & Set Size & Coverage & Set Size & Coverage  & Ave Set Size \\ \midrule
0.12  & 0.12      & 0.06 & 0.950    & 1.581        & 0.942     & 1.483        & 0.987    & 3.353        & 0.976     & 2.009        \\
      &           & 0.12 & 0.968    & 1.839        & 0.959     & 1.671        & 0.996    & 6.731        & 0.986     & 2.387        \\
      &           & 0.18 & 0.985    & 2.761        & 0.974     & 1.946        & 0.999    & 9.417        & 0.990     & 2.666        \\
      &           & 0.25 & 0.997    & 7.078        & 0.985     & 2.369        & 1.000    & 10.000       & 0.994     & 3.213        \\ \cmidrule(l){2-11} 
      & 0.25      & 0.06 & 1.000    & 10.000       & 0.943     & 4.911        & 1.000    & 10.000       & 0.977     & 6.500        \\
      &           & 0.12 & 1.000    & 10.000       & 0.953     & 5.328        & 1.000    & 10.000       & 0.984     & 6.880        \\
      &           & 0.18 & 1.000    & 10.000       & 0.961     & 5.711        & 1.000    & 10.000       & 0.991     & 7.366        \\
      &           & 0.25 & 1.000    & 10.000       & 0.974     & 6.323        & 1.000    & 10.000       & 0.994     & 7.628        \\
      &           & 0.37 & 1.000    & 10.000       & 0.988     & 7.133        & 1.000    & 10.000       & 0.998     & 8.387        \\
      &           & 0.50 & 1.000    & 10.000       & 0.996     & 7.990        & 1.000    & 10.000       & 0.999     & 9.049        \\
      &           & 0.75 & 1.000    & 10.000       & 1.000     & 10.000       & 1.000    & 10.000       & 1.000     & 10.000       \\ \cmidrule(l){2-11} 
      & 0.50      & 0.06 & 1.000    & 10.000       & 0.950     & 8.836        & 1.000    & 10.000       & 0.980     & 9.310        \\
      &           & 0.12 & 1.000    & 10.000       & 0.960     & 8.985        & 1.000    & 10.000       & 0.984     & 9.402        \\
      &           & 0.18 & 1.000    & 10.000       & 0.965     & 9.075        & 1.000    & 10.000       & 0.986     & 9.450        \\
      &           & 0.25 & 1.000    & 10.000       & 0.974     & 9.222        & 1.000    & 10.000       & 0.990     & 9.535        \\
      &           & 0.37 & 1.000    & 10.000       & 0.983     & 9.403        & 1.000    & 10.000       & 0.996     & 9.656        \\
      &           & 0.50 & 1.000    & 10.000       & 0.991     & 9.557        & 1.000    & 10.000       & 0.999     & 9.789        \\
      &           & 0.75 & 1.000    & 10.000       & 0.999     & 9.820        & 1.000    & 10.000       & 1.000     & 9.947        \\ \midrule
0.25  & 0.12      & 0.06 & 0.954    & 2.570        & 0.937     & 2.232        & 0.991    & 7.016        & 0.968     & 2.992        \\
      &           & 0.12 & 0.969    & 3.049        & 0.949     & 2.395        & 0.998    & 9.042        & 0.976     & 3.301        \\
      &           & 0.18 & 0.984    & 4.573        & 0.956     & 2.562        & 1.000    & 9.845        & 0.981     & 3.567        \\
      &           & 0.25 & 0.999    & 9.510        & 0.969     & 2.908        & 1.000    & 10.000       & 0.986     & 3.895        \\ \cmidrule(l){2-11} 
      & 0.25      & 0.06 & 0.953    & 2.051        & 0.941     & 1.836        & 0.984    & 3.307        & 0.974     & 2.551        \\
      &           & 0.12 & 0.960    & 2.183        & 0.950     & 1.951        & 0.991    & 4.077        & 0.981     & 2.832        \\
      &           & 0.18 & 0.969    & 2.411        & 0.959     & 2.126        & 0.994    & 5.242        & 0.984     & 3.054        \\
      &           & 0.25 & 0.979    & 2.790        & 0.969     & 2.394        & 0.997    & 6.749        & 0.988     & 3.295        \\
      &           & 0.37 & 0.991    & 3.867        & 0.981     & 2.858        & 1.000    & 9.660        & 0.994     & 3.888        \\
      &           & 0.50 & 0.999    & 7.824        & 0.989     & 3.480        & 1.000    & 9.948        & 0.996     & 4.564        \\ \cmidrule(l){2-11} 
      & 0.50      & 0.06 & 1.000    & 10.000       & 0.947     & 6.762        & 1.000    & 10.000       & 0.979     & 7.837        \\
      &           & 0.12 & 1.000    & 10.000       & 0.956     & 7.024        & 1.000    & 10.000       & 0.983     & 8.016        \\
      &           & 0.18 & 1.000    & 10.000       & 0.960     & 7.212        & 1.000    & 10.000       & 0.988     & 8.221        \\
      &           & 0.25 & 1.000    & 10.000       & 0.969     & 7.523        & 1.000    & 10.000       & 0.992     & 8.430        \\
      &           & 0.37 & 1.000    & 10.000       & 0.981     & 7.935        & 1.000    & 10.000       & 0.996     & 8.763        \\
      &           & 0.50 & 1.000    & 10.000       & 0.990     & 8.350        & 1.000    & 10.000       & 0.998     & 9.040        \\
      &           & 0.75 & 1.000    & 10.000       & 0.997     & 9.008        & 1.000    & 10.000       & 0.999     & 9.494        \\ \cmidrule(l){2-11} 
0.50  & 0.12      & 0.06 & 0.948    & 3.701        & 0.923     & 3.060        & 0.994    & 8.485        & 0.965     & 4.196        \\
      &           & 0.12 & 0.961    & 4.230        & 0.929     & 3.159        & 0.999    & 9.564        & 0.971     & 4.457        \\
      &           & 0.18 & 0.980    & 5.843        & 0.937     & 3.330        & 1.000    & 9.960        & 0.973     & 4.538        \\
      &           & 0.25 & 0.998    & 9.329        & 0.947     & 3.550        & 1.000    & 10.000       & 0.977     & 4.753        \\ \cmidrule(l){2-11} 
      & 0.25      & 0.06 & 0.943    & 3.152        & 0.925     & 2.792        & 0.990    & 7.254        & 0.969     & 4.013        \\
      &           & 0.12 & 0.951    & 3.417        & 0.933     & 2.929        & 0.995    & 7.818        & 0.973     & 4.172        \\
      &           & 0.18 & 0.961    & 3.771        & 0.942     & 3.112        & 0.996    & 8.476        & 0.976     & 4.276        \\
      &           & 0.25 & 0.970    & 4.095        & 0.948     & 3.246        & 0.998    & 8.916        & 0.978     & 4.416        \\
      &           & 0.37 & 0.987    & 5.773        & 0.959     & 3.586        & 1.000    & 9.958        & 0.984     & 4.753        \\
      &           & 0.50 & 0.999    & 9.121        & 0.970     & 3.952        & 1.000    & 10.000       & 0.988     & 5.171        \\ \cmidrule(l){2-11} 
      & 0.50      & 0.06 & 0.957    & 2.767        & 0.943     & 2.428        & 0.984    & 3.995        & 0.974     & 3.288        \\
      &           & 0.12 & 0.964    & 2.948        & 0.949     & 2.558        & 0.986    & 4.165        & 0.977     & 3.405        \\
      &           & 0.18 & 0.968    & 3.071        & 0.955     & 2.673        & 0.988    & 4.351        & 0.980     & 3.542        \\
      &           & 0.25 & 0.974    & 3.272        & 0.962     & 2.835        & 0.991    & 4.850        & 0.983     & 3.806        \\
      &           & 0.37 & 0.982    & 3.721        & 0.973     & 3.182        & 0.993    & 5.160        & 0.986     & 4.044        \\
      &           & 0.50 & 0.987    & 4.215        & 0.980     & 3.524        & 0.997    & 6.439        & 0.990     & 4.511        \\
      &           & 0.75 & 0.996    & 5.708        & 0.987     & 4.285        & 1.000    & 9.287        & 0.994     & 5.316        \\ \bottomrule
\end{tabular}}
\caption{Comparison of CAS and \method for model trained with various smoothing $\sigma$, and input data with different smoothing $\sigma$. Results are for CIFAR-10 dataset.}
\label{tab:all-sigmas}
\end{table}



\end{document}